\documentclass[runningheads]{llncs}
\usepackage{graphicx}
\usepackage{comment}

\usepackage{amsmath, amssymb, bm, dsfont}
	\DeclareMathOperator{\argmax}{argmax}

\usepackage{algorithm}
\usepackage{algorithmic}
\usepackage{cite}
\usepackage{hyperref}
\usepackage{xr}
\usepackage{xcolor}
\usepackage[bottom]{footmisc}

\begin{document}
\title{Batch Active Learning in Gaussian Process Regression using Derivatives}
%
%
\author{Hon Sum Alec Yu\thanks{The author worked at BCAI when this paper was first written.} 
\and Christoph Zimmer\inst{1} 
\and Duy Nguyen-Tuong\inst{1}
}
%
\institute{Bosch Center for Artificial Intelligence (BCAI), 71272 Renningen, Germany \\ 
\email{}}
\maketitle              
\begin{abstract}
We investigate the use of derivative information for Batch Active Learning in Gaussian Process regression models. The proposed approach employs the predictive covariance matrix for selection of data batches to exploit full correlation of samples. We theoretically analyse our proposed algorithm taking different optimality criteria into consideration and provide empirical comparisons highlighting the advantage of incorporating derivatives information. Our results show the effectiveness of our approach across diverse applications.

\end{abstract}
%
%
\section{Introduction}
\label{sect1}

Active learning (AL) is the process of \textit{sequential} data labelling for supervised learning, while aiming to achieve the desired accuracy with fewer selected training instances possible. This approach is useful in many applications such as image labelling (e.g. \cite{Joshi09multi}) and parameter estimation (e.g. \cite{tong2001active}). Batch AL (BAL), on the other hand, differs by making \textit{parallel} instead of sequential queries, such as in the case of learning height-maps for robot navigation (e.g. \cite{Plagemann2009}) and generation of experiments for controller optimization (e.g. \cite{Schillinger2017sal}). Majority of papers in this field focused on classification problems (e.g. \cite{Guo2008,Chen2013}) and rather few address regression problems (e.g. \cite{Cai2016}). More recently, Schreiter et al.\cite{schreiter2015safe} first applied AL in Gaussian Process (GP) regression and Zimmer et al.\cite{zimmer2018safe} extended the scope to learning time-series models. In learning time-series, they introduced an exploration criterion for active learning based on the determinant of the GP predictive covariance matrix. Inspired by their work, we generalise their idea to the BAL setting for GP regression model. Furthermore, motivated by the insight that in many applications, e.g. learning terrain height-maps from sensor readings, gradient information might help to increase the performance, we also incorporate derivative information into the BAL process.

Derivative information can be straightforwardly incorporated into AL for exploration under a GP framework \cite{rw2006gpml}. Using such technique into GP-based exploration has been done before by several authors in, e.g. Bayesian optimization \cite{wu2017exploiting}. Results from Eriksson et al.\cite{eriksson2018scaling} and Wu et al.\cite{wu2017exploiting} empirically demonstrate the usefulness of derivative information on predictive performance and, to the best of our knowledge, we are the first to provide formal argument  for it and exploiting it in the setting of BAL regression. During exploration, a popular approach is to employ the \textit{entropy criterion} for exploration while using a greedy selection based on the maximum variance of the GP (e.g. \cite{Cohn95,srinivas2012information,schreiter2015safe,zimmer2018safe}). An additional challenge for BAL is that the predictive variance becomes a covariance matrix and one apporach is to apply one of the following optimal designs \cite{settles2009active}: D-optimality for the \textit{determinant}, A-optimality for the \textit{trace} and E-optimality for the \textit{maximum eigenvalue}. Zimmer et al.\cite{zimmer2018safe} provided a theoretical reasoning in using D-optimality in GP for time-series problem. 

In contrast to previous works, we first provide a formal argument in using derivative information by theoretically analyse our proposed algorithm, showing that by defining order of matrices (more in section \ref{sect4}), a ``smaller'' predictive covariance matrix is attained when derivative information is included. We also generalise the work by Zimmer et al.\cite{zimmer2018safe} in analysing the other two optimality criteria, which have been much less studied. We derive a bound on the decay rate of the predictive covariance matrix when derivative information is available. 

To summarise, the main contributions of this paper are:
\begin{itemize}
	\item We formulate a BAL setting with GPs while exploiting derivative information, and evaluate the proposed algorithm on realistic applications.
	\item We provide a theoretical justification of why derivatives help in exploration, and investigate the predictive performance for major optimal criteria.
\end{itemize}

The remaining of this paper is organised as follow: Section \ref{sect2} presents a brief overview on related work. Section \ref{sect3} introduces GP with derivatives and describe our BAL algorithm. Section \ref{sect4} shows our theoretical analysis and section \ref{sect5} evaluates our concept on several test scenarios. A general discussion is given in Section \ref{sect6}. The supplementary materials provide further details on our approach, theoretical proofs and experiments.

\section{Related Work}
\label{sect2}

GPs with derivatives have been studied in statistics and machine learning in the past (e.g. \cite{adler1981geometry,papoulis1991probability,solak2002derivative, rasmussen2003gphmc}) and theoretical properties on kernel derivatives were studied in applied mathematics community (e.g. \cite{buescu2006positive, buescu2006Bpositive,zhou2008derivative, massa2017estimates}). In machine learning community, this field has been considered in several applications over the last decade. For example, Riihim\"{a}ki et al.\cite{riihimaki2010gaussian} applied GPs with derivatives to learn functions with monotonicity information, whereas Zhang et al.\cite{zhang2013learning} employed GP with derivatives to incorporate invariance information, providing some theoretical bounds under specific classes of kernels. Wong et al.\cite{wong2015svd} used derivatives of kernels to improve learning with geometric data. This concept has been generalised in Bayesian Optimisation by Wu et al.\cite{wu2017bayesian}. They defined a specific metric to exploit derivative information and outlined some theoretical results from that metric. Including higher order derivatives into GP has also been investigated by Wu et al.\cite{wu2017exploiting}. Eriksson et al.\cite{eriksson2018scaling} showed empirically that GP with derivatives has a lower predictive variance. 

While the above works showed the advantage of derivative information, we additionally consider AL as a query strategy. AL is itself a broad topic and here we only refer readers to Settles\cite{settles2009active} and Dasgupta\cite{dasgupta2011two} for the in-depth survey of the algorithmic and theoretical ideas. For BAL, this field has also been investigated under different settings in machine learning literature (e.g. \cite{Hoi2006, Guo2008, Chen2013, Cai2016}). While most work focused on classification problems (e.g. \cite{Hoi2006, Guo2008}), relatively few studies have been conducted for regression (e.g. \cite{Cai2016, Wu2016}). Different selection criteria can be employed for selecting batches, such as Fisher information matrix (e.g. \cite{Hoi2006}), matching via Monte Carlo simulation (e.g. \cite{azimi2012batch}) and expected model change (e.g. \cite{Chen2013}). Recently, Pinsler et al.\cite{pinsler2019bayesian} focused on Bayesian BAL, mitigated the issue of selecting the most informative data in large-scale. The main difference between previous works and ours is that we additionally use gradient info for AL.

In the context of BAL in GPs, Houlsby et al.\cite{houlsby2011bayesian} applied Bayesian Active Learning on GP classification problem where they defined an acquisition function which estimates the quantity of mutual information between the model predictions and the model parameters. Kirsch et al.\cite{kirsch2019batchbald} extended their work into the context of BAL. Their focus was to derive a tractable approximation to the mutual information. Low et al.\cite{low2012decentralized} and Ouyang et al.\cite{ouyang2014multi} are another examples of AL in GPs. However, the former focused on classification whereas this paper is on regression. The latter paper focused on designing a specific metric for active learning exploration and they showed that their metric is equivalent to maximising entropy. There are relatively few papers tackling BAL in GP in the context of regression. In safe dynamic exploration, Schreiter et al.\cite{schreiter2015safe} is arguably the first paper to work on AL in GP regression but they only explore a single point at a time. Yu et al. extends this work to state space models \cite{yu2021algpssm} and Li et al. extends to GPs with multiple outputs \cite{li2022multioutputgp}. Zimmer et al.\cite{zimmer2018safe} proposed an approach to select trajectories based on the GP framework. Their idea was to employ a parametrization of the discretised trajectory, where the GP is defined on discretization points. It is intuitive to select D-optimality as the optimality criterion, partly because this refers closely to the log-determinant. This quantity is frequently encountered in GPs literature as well as wider machine learning problems, with literatures dedicated into this matter (e.g. \cite{dong2017scalable, han2015large, fitzsimons2017bayesian}). In contrast to previous works, we formulate an approach generalizing their ideas to the BAL settings and construct the BAL strategy including the other two less studied optimality criteria, namely A-optimality and E-optimality.

\section{Batch Active Learning in GP Regression}
\label{sect3}

We consider a regression problem where $n_0$ $d$-dimensional initial inputs $\mathbf{x}_{1: n_0} \!=\! \lbrace \mathbf{x}_i \!=\! (x_{i, 1}, \!\cdots\!,  x_{i, d}) \rbrace^{n_0}_{i=1} \!\in\! \mathcal{X} \!\subseteq\! \mathbb{R}^{n_0 \times d}$ and outputs $y_{1: n_0} \!=\! \lbrace y_i \rbrace^{n_0}_{i=1} \!\in\! \mathcal{Y} \!\subseteq\! \mathbb{R}^{n_0}$ are given and our goal is to learn the mapping between $\mathcal{X}$ and $\mathcal{Y}$. A GP is a collection of random variables where any finite number of which have a joint multivariate normal distribution specified by its mean function $m \!:\! \mathbb{R}^d \!\rightarrow\! \mathbb{R}$ and covariance function $k \!:\! \mathbb{R}^d \!\times\! \mathbb{R}^d \!\rightarrow\! \mathbb{R}$. For any $\mathbf{x}_i, \mathbf{x}_j \!\in\! \mathcal{X}$, the distribution $f$ is commonly written as $ f \!\sim\! \mathcal{GP}(m(\mathbf{x}_i), k(\mathbf{x}_i, \mathbf{x}_j))$\cite{rw2006gpml}. The prior mean function is often set to zero and the covariance function encodes our assumptions about the function we would like to learn. The distribution of the model with a Gaussian output noise $\sigma^2$ is
\begin{equation} \label{gpr_model}
p\left(y_{1: n_0} \vert \mathbf{x}_{1: n_0} \right) = \mathcal{N}\left( y_{1:n_0} \vert \mathbf{0}, \mathbf{K}_{n_0} + \sigma^2 \mathbf{I}_{n_0} \right),
\end{equation}
where $\mathbf{K}_{n_0} \!=\! \left( k(\mathbf{x}_i, \mathbf{x}_j) \right)^{n_0}_{i, j=1}$ is the covariance matrix and $\mathbf{I}_{n_0}$ is an identity matrix. Due to the need of derivative information, we need the covariance function to be at least twice differentiable. A prime example is the \emph{squared exponential kernel}, which is given by $k( \mathbf{x}_i, \mathbf{x}_j) \!=\! \sigma_{f}^2 \exp(-\tfrac{1}{2}( \mathbf{x}_i \!-\!  \mathbf{x}_j)^T \boldsymbol{\Lambda}_{f}^2( \mathbf{x}_i \!-\!  \mathbf{x}_j))$, parametrized by the hyper-parameters $\boldsymbol{\theta}_f \!=\! (\sigma_{f}^2,\boldsymbol{\Lambda}_{f}^2)$. This kernel function will be applied throughout this paper and the derivation of its derivatives are left in the supplementary materials. We postulate that similar results hold for different twice differentiable kernels.

The predictive distribution of a batch of $n$ outputs $y^*_{1:n} \!\in\! \mathcal{Y}$ given inputs $\mathbf{x}^*_{1:n} \!\in\! \mathcal{X}$ is given by
\begin{equation}\label{gpr_pred}
p\left( y^*_{1:n} \vert \mathbf{x}^*_{1:n}, y_{1: n_0}, \mathbf{x}_{1: n_0} \right) = \mathcal{N} (y^*_{1: n} \vert \bm{\mu}(\mathbf{x}^*_{1: n}), \bm{\Sigma}(\mathbf{x}^*_{1: n})),
\end{equation}
where $\bm{\mu}(\mathbf{x}^*_{1: n})$ and $\bm{\Sigma}(\mathbf{x}^*_{1: n}))$ can be written in closed form as described in section 2.2 in \cite{rw2006gpml}. Note that under BAL, in contrast to a single point prediction, the predictive mean $\bm{\mu}(\mathbf{x}^*_{1: n})$ becomes a $n\!\times\! 1$ vector and the predictive covariance $\bm{\Sigma}(\mathbf{x}^*_{1: n})$ is a $n\! \times\! n$ matrix.

\subsection{Gaussian Process Regression with Derivatives}
\label{sect3_part1_gpd}

The derivative of a GP is another GP because differentiation is a linear operator (section 9.4 in \cite{rw2006gpml}). Suppose derivative information is available given the inputs, denoted as $\nabla y_{1: n_0} $, the distribution of the model with a Gaussian output noise $\sigma^2$ can be written similarly as
\begin{align} \label{gprd_model}
p\left( y_{1: n_0}, \nabla y_{1: n_0} \vert \mathbf{x}_{1: n_0} \right) = \mathcal{N} \left( y_{1: n_0}, \nabla y_{1: n_0} \vert \mathbf{0}, \tilde{\mathbf{K}}_{n_0} + \sigma^2 \mathbf{I}_{n_0 \times (1 + d)} \right).
\end{align}  
$\nabla y_{1: n_0}$ is the collection of the gradient of each point $y_i$ in Equation \ref{gprd_model} for all $n_0$ initial points, so this is a $n_0 d \times 1$ vector. $\tilde{\mathbf{K}}_{n_0}$ is the extended covariance matrix including derivative information. With similar derivation, the corresponding predictive mean and covariance are
\begin{equation} \label{gprD_pred}
p(y^*_{1: n}, \nabla y^*_{1: n} \vert \mathbf{x}^*_{1: n}, y_{1: n_0}, \nabla y_{1: n_0}, \mathbf{x}_{1: n_0}) = \mathcal{N} \left( y^*_{1: n}, \nabla y^*_{1: n} \vert \tilde{\bm{\mu}}(\mathbf{x}^*_{1: n}),  \tilde{\bm{\Sigma}}(\mathbf{x}^*_{1: n}) \right),
\end{equation}
where 
\begin{equation} \label{gprD_predmean}
\tilde{\bm{\mu}}(\mathbf{x}^*_{1: n}) := 
\begin{pmatrix}
\tilde{\bm{\mu}}_p(\mathbf{x}^*_{1: n}) \\
\tilde{\bm{\mu}}_g(\mathbf{x}^*_{1: n})
\end{pmatrix},
\end{equation}
\begin{equation} \label{gprD_predcov}
\begin{aligned}[b]
\tilde{\bm{\Sigma}}(\mathbf{x}^*_{1: n}) &:= 
\begin{pmatrix}
\tilde{\bm{\Sigma}}_p(\mathbf{x}^*_{1: n}) & \tilde{\bm{\Sigma}}_{pg}(\mathbf{x}^*_{1: n}) \\
\tilde{\bm{\Sigma}}_{gp}(\mathbf{x}^*_{1: n}) & \tilde{\bm{\Sigma}}_g(\mathbf{x}^*_{1: n})
\end{pmatrix}.
\end{aligned}
\end{equation}

Note that the new predictive mean becomes a $n (1\!+\! d)\! \times\! 1$ vector, whereas the predictive covariance becomes a $n (1 \!+\! d) \times n (1 \!+\! d)$ matrix. The new predictive distributions, i.e. Equation \ref{gprD_predmean}, \ref{gprD_predcov} are analogous to the case without derivatives except the additional gradient terms. Specifically, we can write $\tilde{\bm{\mu}}_p(\mathbf{x}^*_{1:n})\!\in\!\mathbb{R}^{n}$ as the predictive mean with respect to the point, whereas  $\tilde{\bm{\mu}}_g(\mathbf{x}^*_{1:n})\!\in\!\mathbb{R}^{nd}$ refers to the gradient. Similarly, $\tilde{\mathbf{\Sigma}}_p(\mathbf{x}^*_{1: n}) \!\in\!\mathbb{R}^{n \times n}$ is written as the predictive covariance with respect to the points, $\tilde{\mathbf{\Sigma}}_g(\mathbf{x}^*_{1: n}) \in \mathbb{R}^{dn \times dn}$ is the corresponding covariance with respect to the gradients. $\tilde{\bm{\Sigma}}_{pg}(\mathbf{x}^*_{1: n})$ and $\tilde{\bm{\Sigma}}_{gp}(\mathbf{x}^*_{1: n})$ are crossing terms, which are non-zeros in general but symmetric. For brevity, we leave the details in supplementary materials. 

In practice, there is a question of the availability of derivative information and efforts required to obtain them. It turns out that there are different approaches in obtaining or estimating gradients from data, which are not that laborious. This will be further elaborated in Sections \ref{sect5} \& \ref{sect6}.

\subsection{The Active Learning Algorithm} 
\label{sect3_part2_ALAlgo}

Active learning is a query strategy to pick the most suitable batch of new points $\mathbf{x}^*_{1: n}$ to learn the regression model as we explore. With derivative information, we can have different choice of covariance matrices in doing exploration, e.g. the full matrix $\tilde{\mathbf{\Sigma}}(\cdot)$, or its sub-matrix $\tilde{\mathbf{\Sigma}}_p(\cdot)$. As we intend to find informative points, the sub-matrix $\tilde{\mathbf{\Sigma}}_p(\cdot)$ is a reasonable choice. This matrix corresponds to the predictive covariance with respect to the points and the size of which is $n \!\times\! n$. As it has the same size as the covariance matrix $\mathbf{\Sigma}(\cdot)$, i.e. BAL \textit{without} gradient information, we are able to directly compare these matrices for further investigation. Also, the size of $\tilde{\mathbf{\Sigma}}_p(\cdot)$ is significantly less than the full matrix $\tilde{\mathbf{\Sigma}}(\cdot)$, which reduces the computational demand. 

Therefore, the BAL strategy to obtain a new batch of points $\mathbf{x}^*_{1: n}$ is to solve the optimisation problem
\begin{equation} \label{optNewPt}
\mathbf{x}^*_{1: n} = \argmax_{\mathbf{x}_{1: n} \in \mathcal{X}} \mathcal{I} \left( \tilde{\mathbf{\Sigma}}_p(\mathbf{x}_{1: n}) \right),
\end{equation}
where $\mathcal{I}$ are the three optimal designs (D-optimality, A-optimality or E-optimality) in this paper. It should be noted that even  $\tilde{\mathbf{\Sigma}}_p(\cdot)$ also contains derivative information despite being a sub-matrix (see supplementary materials). In practice, at each round of exploration (up to final round $T$) we can have different batch size, so we denote this as $n_t$ for round $t = 1, \cdots, T$.

The full procedure of BAL using GPs with derivatives is summarized in Algorithm \ref{alGPD} and we abbreviate our approach as \emph{BALGPD}. Since the three optimal designs are continuous functions, equation \ref{optNewPt} can be optimised using gradient-based approaches. $\mathbf{x}_{1:n}$ can be initialized uniformly or via clustering as a pre-processing step for optimisation. A similar approach was employed for optimisation of pseudo-inputs in the sparse GP setting \cite{snelson06}. Note that in regression problem, $y^*_{1: n_t}$ and $\nabla y^*_{1: n_t}$ are continuous observations and we do not have the concept of classes for them. Nevertheless, since we update model hyperparameters at every step after we get the new data and re-train the model, the updates of the hyperparameters implicitly reflect the usage of $y^*_{1: n_t}$ and $\nabla y^*_{1: n_t}$. 
 
\begin{algorithm}[t]
	\caption{Batch Active Leaning using Gaussian Process (Regression) with Derivatives (BALGPD)}\label{alGPD}
	\begin{algorithmic}
		\STATE {\bfseries Input:} Initial training data $\{\mathbf{x}_{1:n_0}, y_{1: n_0}, \nabla y_{1:n_0}\}$, batch size $n_t$ for each round.
		\STATE Train the initial GP model with derivatives
		\FOR {$t = 1$ to $T$}
		\STATE {1.} Sample initial $n_t$ points: $\mathbf{x}_{1: n_t}$.
		\STATE {2.} Optimize Equation \ref{optNewPt} for $\mathbf{x}_{1: n_t}^*$ using, for example, gradient-based approach.
		\STATE {3.} Evaluate $y^*_{1: n_t}$ and $\nabla y^*_{1: n_t}$ at inputs $\mathbf{x}^*_{1: n_t}$
		\STATE {4.} Update the training data set 
		\STATE {5.} Update the GP regression model with derivatives, according to Equation \ref{gprD_pred}. The model is re-trained per step.
		\ENDFOR
		\STATE {\bfseries Output:} The final GP model with derivatives and the data set including the explored batch of points.
	\end{algorithmic}
\end{algorithm}

\section{Theoretical Analysis}
\label{sect4}

In this section, we present theoretical results from our approach. By looking into the concept of \emph{Information Gain}, we first discuss how derivatives can provide us extra information. Then, we show theoretical guarantees based on the use of $\tilde{\mathbf{\Sigma}}_p(\cdot)$ and investigate the convergence rate when using this covariance matrix for exploration. As stated in Section \ref{sect3}, we assume using the \emph{squared exponential kernel}. For simplicity, we assume the same batch number is selected for all rounds of exploration. That is, $n_t \equiv n$ for all $t$.

\subsection{Information Gain}
\label{sect41}

\emph{BALGPD} includes derivative information and we expect our approach to outperform the same layout but without derivative information (\emph{BALGP}). We first investigate if this claim can be verified mathematically. To begin, we recap the definition of \emph{Information Gain}:

\begin{definition}\label{def_IG}
The \emph{Information Gain} between the functional and observations can be defined as $IG(\mathbf{x}_{1:n}) := I(\lbrace y^*_{1: n, \tau} \rbrace^t_{\tau=1} ;\lbrace \mathbf{f}_{1: n, \tau} \rbrace^t_{\tau=1})$, where $\mathbf{f}_{1: n}$ is a condensed notation of $\lbrace f(\mathbf{x}_1), \cdots,  f(\mathbf{x}_n) \rbrace$.

In the case of GP where the mapping between the two sets is a Gaussian distribution \cite{cover2006elements}, the Information Gain after observing $t$ rounds, with a total of $N_t (= nt)$ points observed, is given by
\begin{equation}\label{def_InfoGain_GP_eq}
IG(\mathbf{x}_{1:n}) = \frac{1}{2} \log \det \left( \mathbf{I} + \sigma^{-2} \mathbf{K}(\mathbf{x}_{1:n}) \right).
\end{equation}

The \emph{Maximum Information Gain} is often written as
\begin{eqnarray}\label{def_MaxInfoGain_GP_eq}
\gamma_t := \max_{\mathcal{A} \subseteq \mathcal{X}; \vert \mathcal{A} \vert = N_t} IG(\mathcal{A})
\end{eqnarray}
\end{definition} 

\begin{remark}
The maximum information gain is in general different depending on whether we are in the learning scheme with derivative information or not. Throughout this section, we use tilde at the top to differentiate between them. That is, when the data collection and learning scheme goes with derivatives information, the maximum information gain will be denoted as $\tilde{\gamma}_t$. The matrix $\mathbf{K}$ becomes $\mathbf{K}_{N_t}$ for \emph{BALGP} or $\tilde{\mathbf{K}}_{N_t}$ for \emph{BALGPD}. The size of the identity matrix $\mathbf{I}$ is adjusted accordingly. 
\end{remark}

With the definition we can show directly that the \emph{BALGPD} process more information than \emph{BALGP}, formalised by the following statement:
\begin{proposition}\label{prop_ExtraInfoGain}
Given the same set of points $\mathbf{x}_{1:n}$, the Information Gain under \emph{BALGPD} is no less than \emph{BALGP}.
\end{proposition}

\begin{proof}[Sketch]
Direct computation using determinant and eigenvalues inequalities. See supplementary materials.
\end{proof}

It is also possible to compute the upper bound of the maximum information gain $\tilde{\gamma}_t$ under \emph{BALGPD}. To begin, we recall from Mercer's theorem \cite{rw2006gpml} that given a positive semidefinite kernel $k(\cdot, \cdot)$ with a finite measure, the kernel can be expressed in terms of the sum of its orthonormal eigenfunctions $\phi_s(\cdot)$ and their associated eigenvalues $\lambda_s \geq 0$ (this sum can be finite or infinite and without loss of generality, $\lambda_1 \geq \lambda_2 \geq \cdots$). That is, for $\mathbf{x}_i, \mathbf{x}_j \in \mathcal{X}$,
\begin{equation}\label{Mercer}
k(\mathbf{x}_i, \mathbf{x}_j) = \sum_{s \geq 0} \lambda_s \phi_s(\mathbf{x}_i) \phi_s(\mathbf{x}_j).
\end{equation}

Then, $\tilde{\gamma}_t$ under \emph{BALGPD} can be upper bounded by these eigen-expansions. Specifically:
\begin{proposition}\label{prop_UpperBoundInfoGain_GPD}
For $\mathbf{x}_i \in \mathcal{X}$ and suppose $\tilde{\mathbf{K}}_{N_t}$ has the eigenexpansion as in Equation \ref{Mercer}. Then,
\begin{equation} \label{prop_UpperBoundInfoGain_eq1}
\tilde{\gamma}_t^{BALGPD} \leq 
\frac{k}{2} \sum_{s \geq 1} \log \left( 1 + \sigma^{-2} \lambda_s \sum^{N_t}_{i=1} (\tilde{\phi}_s(\mathbf{x}_i))^2 \right),
\end{equation}
where $k = (1 - \exp(-1))^{-1}$ and 
\[ (\tilde{\phi}_s(\mathbf{x}_i))^2 := \left( (\phi_s(\mathbf{x}_i))^2 + \sum^d_{j=1}\left( \frac{\partial}{\partial x_{i, j}} (\phi_s(\mathbf{x}_i)) \right)^2 \right). \]
\end{proposition}

\begin{proof}[Sketch.]
The proof extends the idea of Lemma 1 in Seeger et al.\cite{seeger2008information} and Lemma 7.6 in Srinivas et al.\cite{srinivas2012information}. Details in supplementary materials.
\end{proof}

\begin{remark}
A similar upper bound for $\tilde{\gamma}_t$ under \emph{BALGP} can be shown and this is indeed the same as Proposition \ref{prop_UpperBoundInfoGain_GPD} except $\tilde{\phi}_s(\mathbf{x}_i) = \phi_s(\mathbf{x}_i)$.
\end{remark}

\subsection{Decay in Predictive Covariance Matrix}
\label{sect42}

Predictive variance of a GP model should decay eventually as more data are explored and we would like to show that the predictive covariance matrix under \emph{BALGPD} ($\tilde{\mathbf{\Sigma}}_p$) is lower than that under \emph{BALGP} ($\mathbf{\Sigma}$) and they both decay at a specific rate. Since we are using covariance matrices instead of variances, we first need a notion of order of matrices. Fortunately, since both matrices are of the same size, symmetric and positive definite, we can apply the concept of \emph{L\"{o}wner partial ordering} to show the following statement:
\begin{theorem} \label{thm_predCov}
Given the same space to explore and a selected batch of points $\mathbf{x}_{1:n} \in X$, the predictive covariance matrix computed via \emph{BALGPD} is always upper bounded by that via \emph{BALGP} in the sense of L\"{o}wner partial ordering. That is,
\begin{equation}\label{thm_predCov_eq}
\mathbf{\Sigma}(\mathbf{x}_{1:n}) \succeq \tilde{\mathbf{\Sigma}}_p(\mathbf{x}_{1:n}).
\end{equation}
\end{theorem}
\begin{proof}[Sketch]
This is essentially an application of the variational characterisation of Schur complement\cite{lami2016schur}. Details in supplementary materials.
\end{proof}

In the case where batch number is 1, the predictive covariance reduces to variance (i.e. a positive real number) and the above statement reduces to an inequality. Theorem \ref{thm_predCov} extends from a simple inequality to matrix inequality. Indeed, under the three optimality criterion (determinant, maximal eigenvalue and trace), the quantity computed from \emph{BALGPD} is always smaller than that from \emph{BALGP} (more in supplementary materials). 

We can further show that under the three optimality criterion, the predictive covariance decays in a specific rate and this is given by the following statement:
\begin{theorem} \label{thm_convergence_tr}
Collected under the scheme with derivatives information, let $\lbrace \tilde{\mathbf{x}}_{1: n, \tau} \rbrace^t_{\tau=1}$ be $t$ arbitrary explored batch of points within a compact and convex domain and $k$ is a \emph{squared exponential kernel} with $k(\cdot, \cdot)\leq 1$. Let $\tilde{\mathbf{\Sigma}}_{p, \tau-1}(\mathbf{x}_{1:n, \tau})$ be the predictive covariance matrix with derivative information used at step $\tau$ coming from the exploration scheme in equation \ref{optNewPt} (\emph{BALGPD}), then, 
\begin{equation} \label{thm_convergence_tr_eq}
\frac{1}{t} \sum^t_{\tau=1} \mathcal{I}(\tilde{\mathbf{\Sigma}}_{p, \tau-1}(\tilde{\mathbf{x}}_{1:n, \tau})) = \mathcal{O}\left(\frac{\log(t)^{d+1}}{t} \right).
\end{equation},
where $\mathcal{I}$ can be determinant, trace or maximum eigenvalue.
\end{theorem}
\begin{proof}[Sketch]
The proof in the case of \emph{determinant} follows from theorem \ref{thm_predCov}, as well as theorem 2 in Zimmer et al.\cite{zimmer2018safe}. However, further properties are required such that the \emph{trace} of such matrix process the same decay rate. The main idea is to first compute the decay rate under \emph{BALGP}, the decay rate under \emph{BALGPD} can again be computed using theorem \ref{thm_predCov}. For the case of \emph{maximum eigenvalue}, we make use of the fact that trace is the sum of all eigenvalues. Details in supplementary materials.
\end{proof}

Therefore, with the same decay rate, we can freely choose between determinant, trace and maximum eigenvalue.

\section{Empirical Evaluation}
\label{sect5}

In this section, we present several experiments to highlight the advantage and diverse usages of \emph{BALGPD}. In order to visualise our aproach, we first conduct an experiment on a simulated function. Then, we apply our algorithm to a real industrial setting and geographical data, with remarks on practical challenges and solutions. All experiments were implemented based on \texttt{GPyTorch} by Gardner et al.\cite{gardner2018gpytorch}.

\subsection{Simulated Function} 
\label{sect5_exp1}

We review a 2D-cardinal sine function studied in \cite{schreiter2015safe}. In this experiment, we assume a small noise $\pm 0.01$ on both point and derivative measurements and set a fixed batch $n=2$ for each round of exploration. We begin with 4 initial points picked randomly between the range $[-0.5, 0.5]$ and explore the function between $[-10, 15]$. Details of the function is presented in supplementary materials. 

\begin{figure*}[t!]
	\centering
	\includegraphics[scale=.22]{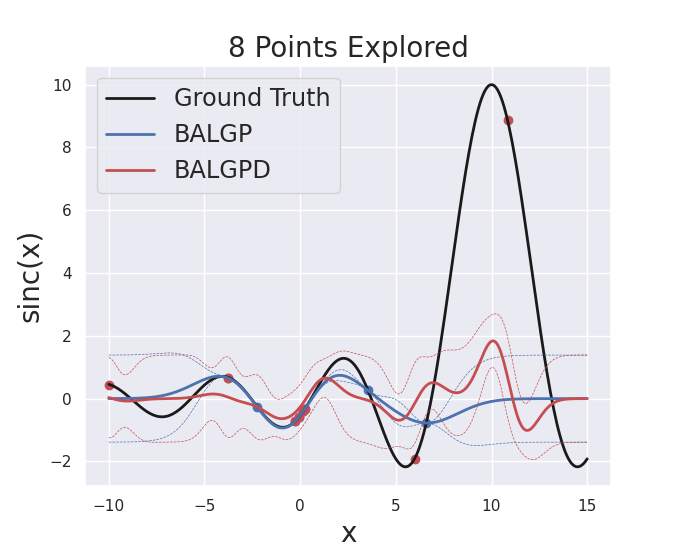}
	\includegraphics[scale=.22]{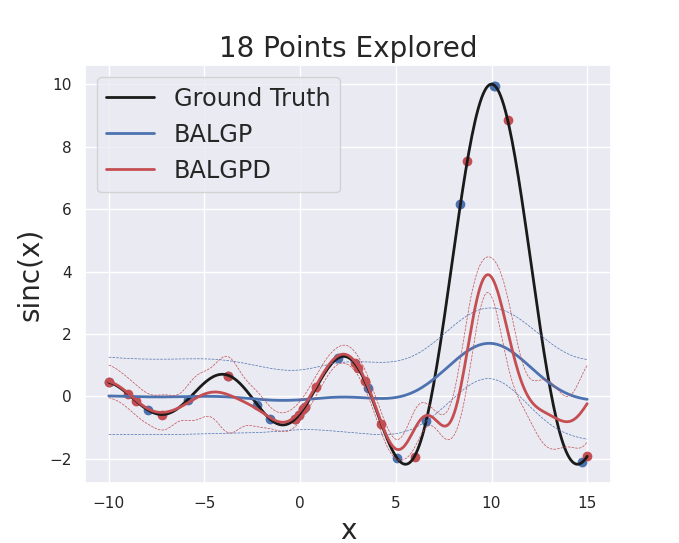}
	\includegraphics[scale=.22]{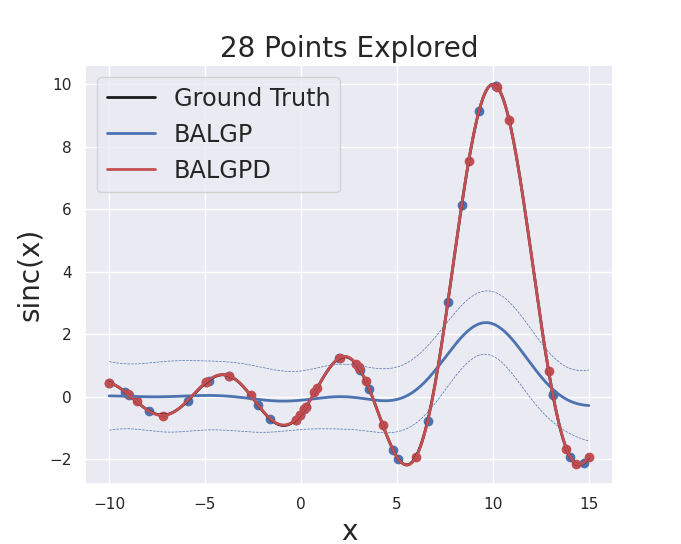}
	\caption{A typical exploration behaviour under \emph{BALGPD} (red) and \emph{BALGP} (blue) for the 2D-cardinal sine function. The diagrams, from left to right, present how the two GP models attain after 8, 18 and 28 points are explored. This sample is taken under A-optimality but the diagrams are similar for other optimality criteria.}
	\label{diagram_exp1_behavior}
\end{figure*}

To visualise our approach, we first pick a sample trial to illustrate a typical behavior of \emph{BALGPD}, presented in figure \ref{diagram_exp1_behavior}. Between \emph{BALGP} and \emph{BALGPD}, we observe that the latter tends to capture regions closer to the turning position (i.e. positions where gradient is about to change sign), especially when not many points have been explored (e.g., when only 8 points are explored). Since this function is rather elementary, when there are more points explored, the visual difference becomes less apparent. Nevertheless, we observe that \emph{BALGPD} visually manage to capture the whole function after 28 points explored, but not \emph{BALGP}. 

\begin{figure*}[t!] 
	\centering
	\includegraphics[scale=.22]{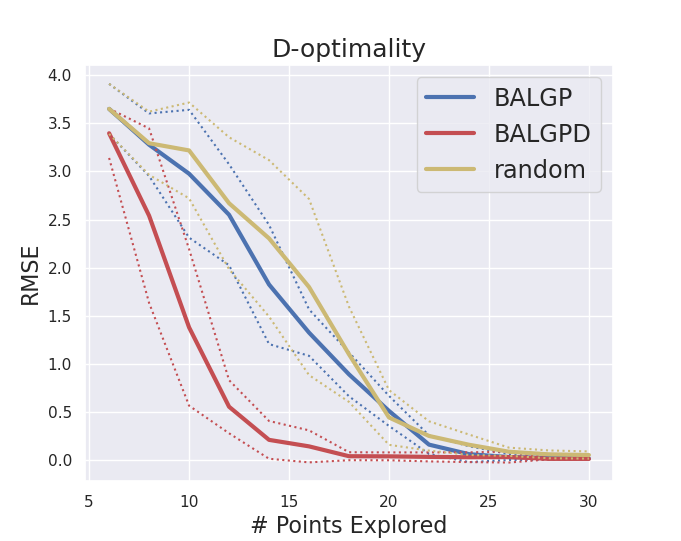}
	\includegraphics[scale=.22]{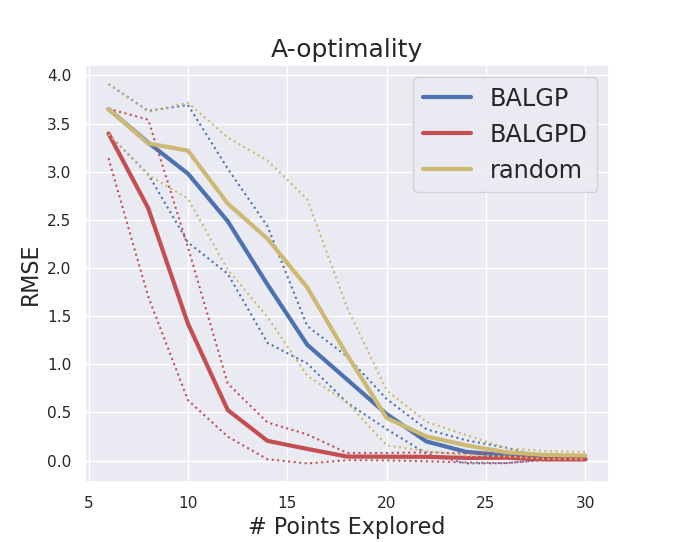}
	\includegraphics[scale=.22]{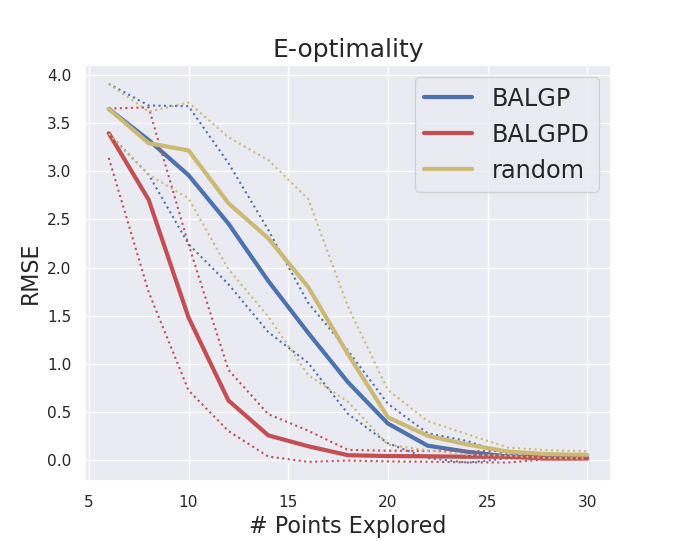}
	\caption{\emph{Simulated function}: The three diagrams present the RMSE attained under \emph{BALGPD} (red), \emph{BALGP} (blue) and no BAL (yellow), as more points are explored. Results are based on D-(left), A-(middle) and E-optimality (right) respectively. The experiment was repeated 30 times.}
	\label{diagram_exp1_results}
\end{figure*}

On the model's predictive performance, results are summarised in Figure \ref{diagram_exp1_results} where \emph{BALGPD} is plotted in red. In order to show that BAL is useful by itself, in addition to the ground case \emph{BALGP} (blue), we also include the results based on random exploration, abbreviated as \emph{random} (yellow). That is, exploration without BAL strategy. To verify our last statement in the previous section that we can freely choose between D-, A- or E-optimality as our optimality criterion, we illustrate the results under all three optimality criterion. It turns out that they all show similar behaviour and \emph{BALGPD} attains a better performance. Some further diagrams and discussion are in supplementary materials.

\subsection{High Pressure Fuel Supply System}
\label{sect5_exp2}

\begin{figure}[h!]
	\centering
	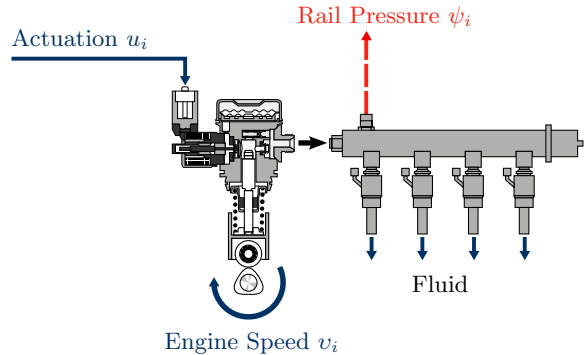
	\caption{A demonstration of a High Pressure Fuel Injection System. At step $i$, the Rail-Pressure (output) $\psi_i$ is given by the Actuation $u_i$ and Engine Speed $\upsilon_i$ at current time-step $i$ and some steps before.} 
	\label{diagram_exp2_figure}
\end{figure}

As a realistic technical use case, we apply \emph{BALGPD} to a high pressure fuel supply system of a modern gasoline in industrial settings. At time $i$, such system gives pressure $\psi_i$ based on the actuation $u_i$ and engine speed $\upsilon_i$. There are different variants but we present a sample of such system, based on \cite{tietze2014model,zimmer2018safe}, in Figure \ref{diagram_exp2_figure}. This is the one which we run simulations to test our approach.

Unlike the simulated function in the previous experiment, the model is now dynamic and employs a non-linear exogenous structure as input. We model $\mathbf{x}_i = (u_i, u_{i-1}, u_{i-2}, u_{i-3}, \upsilon_i, \upsilon_{i-1}, \upsilon_{i-3})$ and $y_i = \psi_i$. Since an inappropriate combination of inputs could result in a hazardously high rail pressure which could damage the physical system, we need to ensure that a safety constraint must be satisfied as we explore. To this end, a safety value $z_i$ will be computed in addition to the output Rail Pressure $y_i$ for a given $\mathbf{x}_i$. To learn the mapping between inputs and safety values, another GP model $g \sim \mathcal{GP}(m_g(\mathbf{x}_i), k_g(\mathbf{x}_i, \mathbf{x}_j))$ is employed to model the safety function. We make use of the safety model $\zeta(\mathbf{x}_{1:n})$ developed by Zimmer et al.\cite{zimmer2018safe}, which maps input points to a safety value and define $\alpha \in (0, 1]$ to be the threshold for considering the input unsafe. To this end, the scope of this experiment is extended to a \emph{Safe BAL} and the optimisation problem becomes
\begin{equation} \label{optNewPt_Safe}
\begin{aligned}
\mathbf{x}^*_{1: n} = \argmax_{\mathbf{x}_{1: n} \in \mathcal{X}} \mathcal{I} \left( \tilde{\mathbf{\Sigma}}_p(\mathbf{x}_{1: n}) \right), \\
\mbox{ such that } \zeta(\mathbf{x}_{1:n}) > 1 - \alpha. 
\end{aligned}
\end{equation}

\emph{BALGPD} needs to be modified in order to accommodate the safety constraint and we present the corresponding algorithm in supplementary materials. We again compare our approach against \emph{BALGP} and random exploration. Note that results for D-optimality under \emph{BALGP} corresponds to the run in Zimmer et al.\cite{zimmer2018safe}. Our aim is to show that \emph{BALGPD} can be extended with safety constraints and still provide a competitive performance.

A practical challenge for this experiment is that only point estimation of the system is possible and we do not know the ground truth function. Therefore, we estimate the gradient for each point via finite differences. In practice, input values in different time are measured and recorded to the physical system and derivatives can be given as part of the outputs in addition to the rail pressure from the device directly.

\begin{figure*}[t!]
	\centering
	\includegraphics[scale=.22]{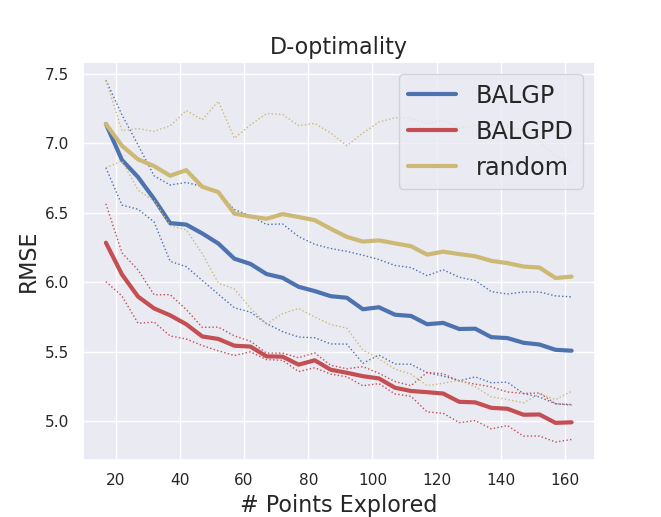}
	\includegraphics[scale=.22]{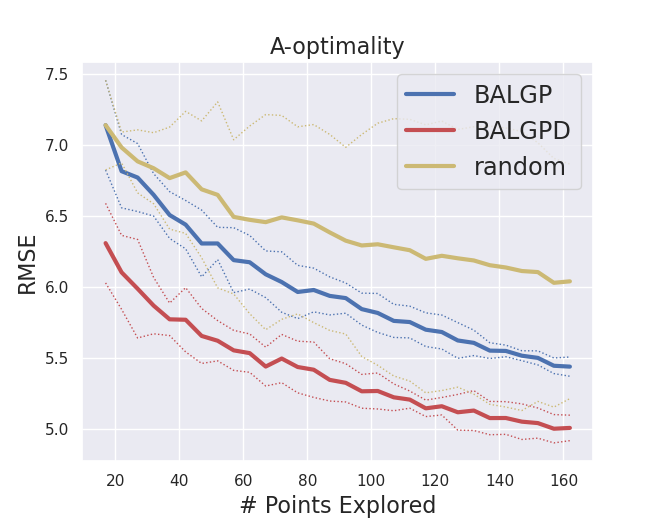}
	\includegraphics[scale=.22]{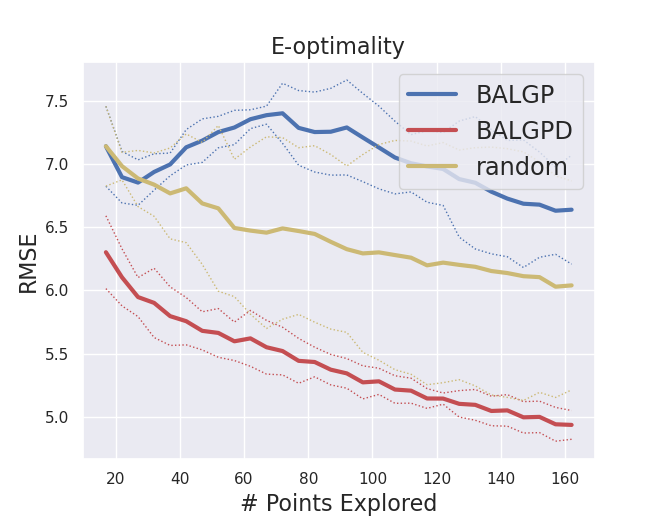}
	\caption{\emph{High Pressure Fuel Supply System}: The three diagrams plot the RMSE attained under \emph{BALGPD} (red), \emph{BALGP} (blue) and random exploration (yellow) as more points are explored, for all D-(left), A-(middle) and E-optimality (right). The experiment was repeated 40 times. Further discussion in suppplementary materials.}
	\label{diagram_exp2_det_tr_maxeig}
\end{figure*}

We maximise the predictive covariance matrix via the optimal \emph{actuation} $u_i$ and \emph{engine Speed} $v_i$. The predictive covariance is computed for a trajectory $\mathbf{x}_{1: n} $, consisting of actuations and engine speeds in different time steps. Note that the trajectory is discretized and, thus, the problem can be formulated as a BAL problem. To begin, we initialise the model with only 5 trajectories in a small safe region told by a domain expert. Each round we pick another trajectory with 5 discretisation points and we explore the space such that the system needs to be safe. We set the threshold $\alpha = 0.5$ (That is, we allow 50\% of the exploration being unsafe. The less the percentage the more cautious the search). Results from running with \emph{BALGPD}, \emph{BALGP} and randomised exploration are presented in Figure \ref{diagram_exp2_det_tr_maxeig}. We show that \emph{BALGPD} does outperform, showing that derivative information are especially useful when the starting data is scarce. 

\subsection{Map Reconstruction} 
\label{sect5_exp3}

In this experiment, we aim to reconstruct a landscape map from data, which is a common problem in geostatistics where we would like to construct a landscape or thematic map (e.g. in geology and robotics) given some positional information. It is very expensive and time-consuming to collect every positional data and as a result, part of positions are selected and a model is used to estimate the surface map. The aim of this experiment is to actively learn the map.

\begin{figure}[h]
	\centering
	\includegraphics[width=8cm]{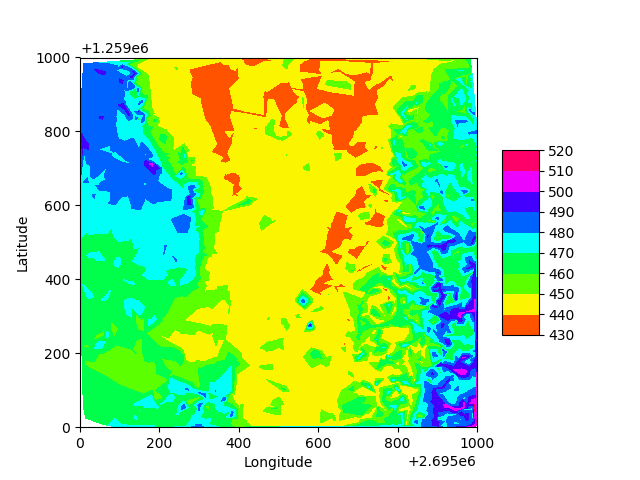}
	\caption{Ground truth of the sub-data, from \texttt{https://shop.swisstopo.admin.ch/en/}.}
	\label{diagram_exp3_surface_GT}
\end{figure}

We pick a precise digital elevation model where data was delivered as a grid with various aperture width. These height models consists of positional coordinates $\mathbf{x} = (x_1, x_2)$ with a height coordinate $y$. The problem is formulated as follow: We assume a tiny portion of labelled positional points to begin and we have a pool of candidate points to explore. We attempt to pick a batch of new positional points for measuring the heights such that our approach (\emph{BALGPD}) can efficiently learn the actual elevation model. A ground-truth of the map studied and its source are stated in supplementary materials. As we only have scattered data in this experiment, finite difference is not possible. We employed an approach from Meyer et al.\cite{meyer2001gradient} for the gradient estimation (this technique is further discussed in supplementary materials). For efficient running, we randomly select a portion from the full data set, i.e. roughly 0.01\% of the full data, which has more than $10^8$ data points. A ground-truth of the such a map is given in Figure \ref{diagram_exp3_surface_GT}.

\begin{figure*}[t!]
	\centering
	\includegraphics[scale=.22]{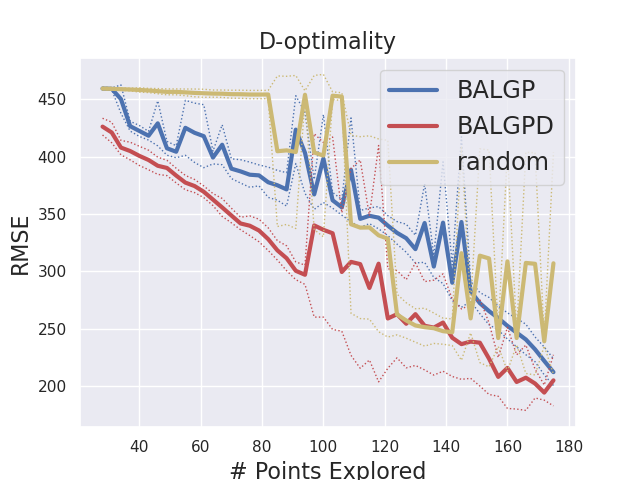}
	\includegraphics[scale=.22]{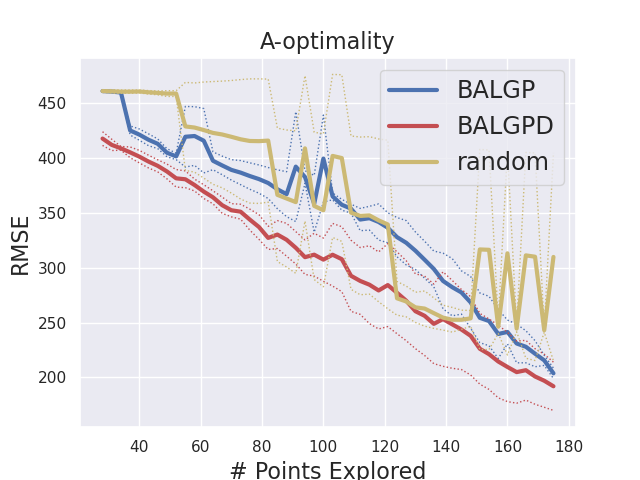}
	\includegraphics[scale=.22]{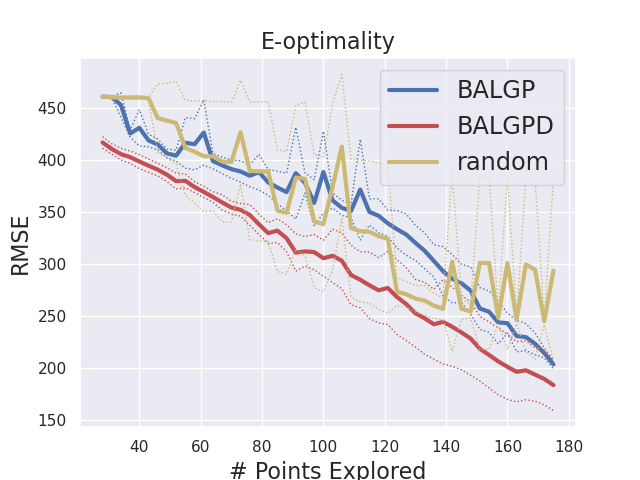}
	\caption{\emph{Map Reconstruction}: The two diagrams plot the RMSE attained under \emph{BALGPD} (red), \emph{BALGP} (blue) and random exploration (yellow) as more points are explored, for all D-(left), A-(middle) and E-optimality (right). The experiment was repeated 60 times. Further discussion in supplementary materials.}
	\label{diagram_exp3_trANDmaxeig}
\end{figure*}

\begin{figure*}[t!]
	\centering
	\includegraphics[scale=.22]{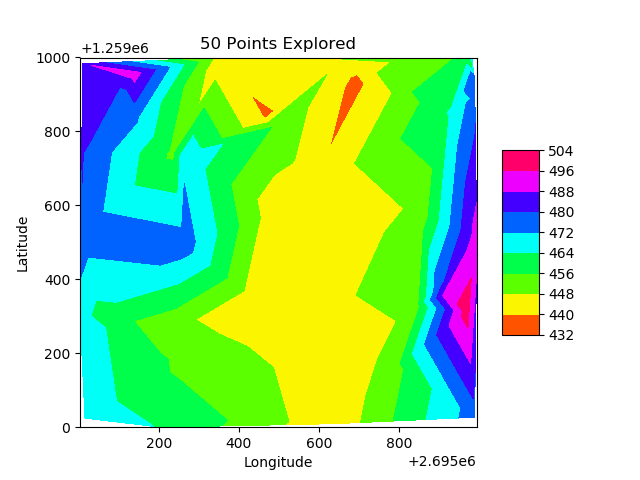}
	\includegraphics[scale=.22]{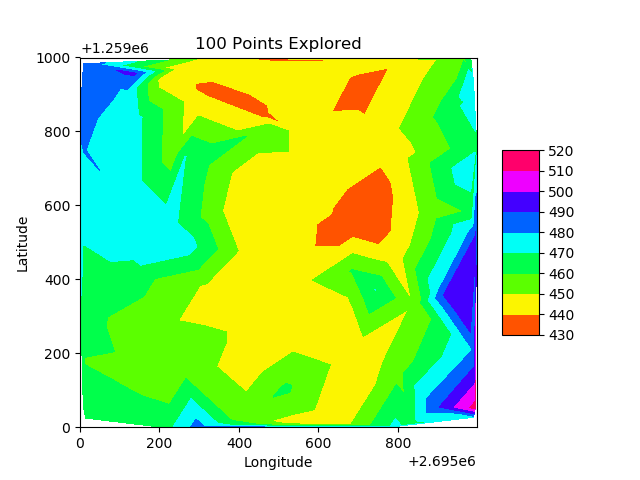}
	\includegraphics[scale=.22]{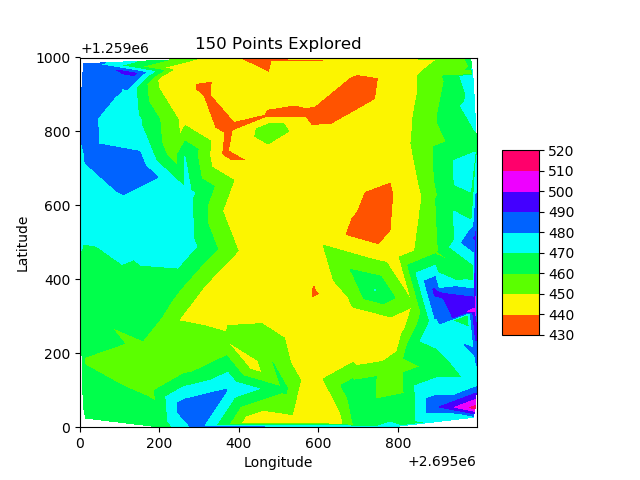}
	\includegraphics[scale=.22]{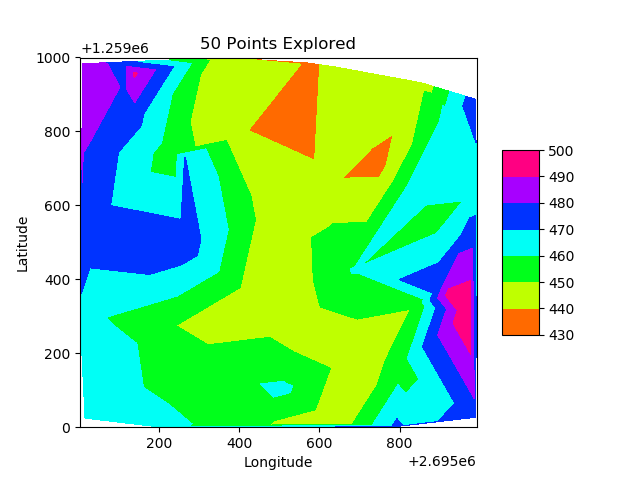}
	\includegraphics[scale=.22]{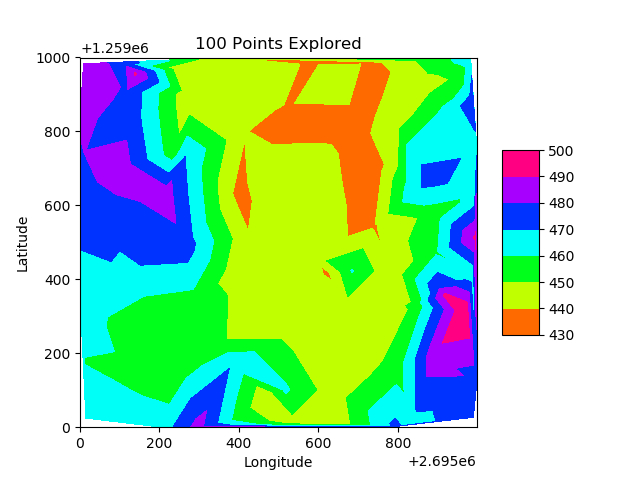}
	\includegraphics[scale=.22]{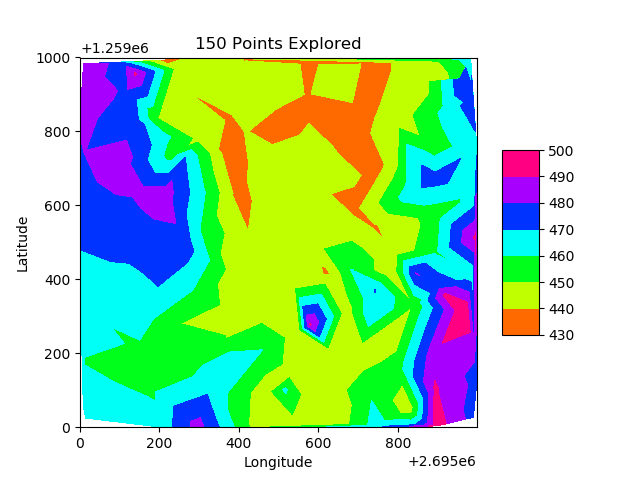}
	\caption{These diagrams show the surface attained from the model when 50 (first column), 100 (second column) and 150 (third column) points are explored. Top row is for \emph{BALGP} and bottom row is for \emph{BALGPD}. It can be seen that with only 150 additional points, i.e. measurements, the reconstructed map using \emph{BALGPD} reflects the main characteristics of the ground-truth.}
	\label{diagram_exp3_surface_model}
\end{figure*}

We assume to have 1\% of the sub-data to begin with and each round we pick a batch of 3 new positional points. Results are presented in Figure \ref{diagram_exp3_trANDmaxeig}. For simpliciaty we only compare between \emph{BALGP} and \emph{BALGPD}. We show that the RMSE attained under \emph{BALGPD} is better. We also present visual behaviour of our approach on this surface map in figure \ref{diagram_exp3_surface_model} to demonstrate how the map is eventually learned. 

\section{Discussion \& Conclusion}
\label{sect6}

In this study, we investigate the effect of including derivatives information into the BAL process using the well-known GP framework. \emph{BALGPD} is theoretically analysed motivating the usage of derivatives for BAL. In particular, we demonstrate their advantages in several realistic scenarios under the three common optimality criteria. While we only present experiments on mechanical systems and geology, we expect our approach to be applicable in other fields with fine practical adjustments, such as determining the structure and dynamics of genetics networks (e.g. \cite{raissi2017machine}), modelling aircraft take-off weights (e.g. \cite{chati2018modeling}) or other autoregressive models in engineering and economics. Also, while current work uses a squared exponential kernel for the experiments, possible future works could incorporate efficient kernel selection strategies (e.g. \cite{bitzer2022kernelbo}) or amortized inference aspects (e.g. \cite{bitzer2023aigph, liu2020aigph}).

%
%
%
\bibliographystyle{splncs04}
\bibliography{reference}

\end{document}


\title{Batch Active Learning in Gaussian Process Regression using Derivatives - Supplementary}
\titlerunning{Batch AL in GPR using Derivatives}
%
\author{Hon Sum Alec Yu\thanks{The author worked at BCAI when this paper was first written.} 
\and Christoph Zimmer\inst{1} 
\and Duy Nguyen-Tuong\inst{1}
}
%
%
\institute{Bosch Center for Artificial Intelligence (BCAI), 71272 Renningen, Germany \\ 
\email{}}
\maketitle              
%

%
%
%

\section{Details on Batch Active Learning in GP Regression}\label{supplementary_BAL}

\subsection{Derivatives of Squared Exponential Kernel} 
\label{supplementary_RBFKernel}

For any $\mathbf{x}_i, \mathbf{x}_j \in \mathcal{X} \subseteq \mathbb{R}^d$, the \emph{squared exponential kernel} is given by
\begin{equation}\label{RBF_kernel}
k(\mathbf{x}_i, \mathbf{x}_j) = \sigma^2_f \exp\left( -\frac{1}{2} (\mathbf{x}_i - \mathbf{x}_j)^T \Lambda^{-1} (\mathbf{x}_i - \mathbf{x}_j) \right),
\end{equation} 
where $\sigma^2_f$ is the kernel variance and $\Lambda$ is the length scale. In this paper, we assume $\Lambda$ as a diagonal matrix with $d$ different elements. That is, $\Lambda = \diag\left(l^2_1, \cdots, l^2_d \right)$. This is a prime example where GP derivatives can be applied and the first and second partial derivatives of this kernel are given by
\begin{align*}
\frac{\partial}{\partial \mathbf{x}_i} k(\mathbf{x}_i, \mathbf{x}_j) &= - \Lambda^{-1} (\mathbf{x}_i - \mathbf{x}_j) k(\mathbf{x}_i, \mathbf{x}_j),  \\
\frac{\partial}{\partial \mathbf{x}_j} k(\mathbf{x}_i, \mathbf{x}_j) &= \Lambda^{-1} (\mathbf{x}_i - \mathbf{x}_j) k(\mathbf{x}_i, \mathbf{x}_j),  \\
\frac{\partial^2}{\partial \mathbf{x}_i \partial \mathbf{x}_j} k(\mathbf{x}_i, \mathbf{x}_j) &= \Lambda^{-1} \left( \mathbf{I}_d - (\mathbf{x}_i - \mathbf{x}_j) (\mathbf{x}_i - \mathbf{x}_j)^T \Lambda^{-1} \right) k(\mathbf{x}_i, \mathbf{x}_j).
\end{align*}
For any differentiable kernel, the first derivative is a $d \times 1$ vector whereas the second derivative is a $d \times d$ matrix.

\subsection{Gaussian Processes with Derivatives - full expression} 
\label{supplementary_GPD}

Recall that the full distribution of the model with a Gaussian output noise $\sigma^2$ is given by
\begin{equation*}
p\left( y_{1: n_0}, \nabla y_{1: n_0} \vert \mathbf{x}_{1: n_0} \right) = \mathcal{N} \left( y_{1: n_0}, \nabla y_{1: n_0} \vert \mathbf{0}, \tilde{\mathbf{K}}_{n_0} + \sigma^2 \mathbf{I}_{n_0 \times (1 + d)} \right).
\end{equation*}  
The gradient of each point $y_i$ in Equation 3 in the main text as denoted as $\nabla y_i$. This is a $d \times 1$ vector as there are $d$ partial derivatives available for  $\mathbf{x}$. We collect the gradients of all $n_0$ initial points as $\nabla y_{1: n_0}$ and this becomes a $n_0 d \times 1$ vector. $\tilde{\mathbf{K}}_{n_0} := \left( \tilde{k}(\mathbf{x}_i, \mathbf{x}_j) \right)^{n_0}_{i, j = 1}$ and each element can be written as
\begin{equation*}
\tilde{k}(\mathbf{x}_i, \mathbf{x}_j) = 
\begin{pmatrix}
k(\mathbf{x}_i, \mathbf{x}_j) & \left( \frac{\partial}{\partial \mathbf{x}_j} k(\mathbf{x}_i, \mathbf{x}_j) \right)^T \\
\frac{\partial}{\partial \mathbf{x}_i} k(\mathbf{x}_i, \mathbf{x}_j) & \frac{\partial^2}{\partial \mathbf{x}_i \partial \mathbf{x}_j} k(\mathbf{x}_i, \mathbf{x}_j)
\end{pmatrix}.
\end{equation*}
While the index goes only $n_0$, the size of this new covariance matrix $\tilde{\mathbf{K}}_{n_0}$ is $n_0(1 + d)$. In practice, the terms are written in blocks and this is denoted by,
\begin{equation*}
\tilde{\mathbf{K}}_{n_0} := 
\begin{pmatrix}
\mathbf{K}_{n_0} & \left( \frac{\partial}{\partial \mathbf{x}_j} \mathbf{K}_{n_0} \right)^T \\
\frac{\partial}{\partial \mathbf{x}_i} \mathbf{K}_{n_0} & \frac{\partial^2}{\partial \mathbf{x}_i \partial \mathbf{x}_j} \mathbf{K}_{n_0}
\end{pmatrix}.
\end{equation*}
We can write the predictive mean and covariance similarly in closed form. When we are given a batch on $n$ new inputs and would like to predict the outputs, the predictive distribution with derivatives information are given by
\begin{equation} \label{gprD_predmeanFULL}
\tilde{\bm{\mu}}(\mathbf{x}^*_{1: n}) = 
\begin{pmatrix}
\tilde{\bm{\mu}}_p(\mathbf{x}^*_{1: n}) \\
\tilde{\bm{\mu}}_g(\mathbf{x}^*_{1: n})
\end{pmatrix} = 
\tilde{k}(\mathbf{x}_{1: n_0}, \mathbf{x}^*_{1: n})^T \left( \tilde{\mathbf{K}}_{n_0} + \sigma^2 \mathbf{I}_{n_0(1 + d)} \right)^{-1} 
\begin{pmatrix}
y_{1: n_0} \\
\nabla y_{1: n_0}
\end{pmatrix}
\end{equation}

\begin{equation} \label{gprD_predcovFULL}
\begin{aligned}[b]
\tilde{\bm{\Sigma}}(\mathbf{x}^*_{1: n}) &:= 
\begin{pmatrix}
\tilde{\bm{\Sigma}}_p(\mathbf{x}^*_{1: n}) & \tilde{\bm{\Sigma}}_{pg}(\mathbf{x}^*_{1: n}) \\
\tilde{\bm{\Sigma}}_{gp}(\mathbf{x}^*_{1: n}) & \tilde{\bm{\Sigma}}_g(\mathbf{x}^*_{1: n}).
\end{pmatrix} \\
&= \tilde{k}(\mathbf{x}^*_{1: n}, \mathbf{x}^*_{1: n}) - \tilde{k}(\mathbf{x}_{1: n_0}, \mathbf{x}^*_{1: n})^T \left(\tilde{\mathbf{K}}_{n_0} + \sigma^2 \mathbf{I}_{n_0(1 + d)}\right)^{-1} \tilde{k}(\mathbf{x}_{1: n_0}, \mathbf{x}^*_{1: n}).
\end{aligned}
\end{equation}
$\tilde{\bm{\Sigma}}_{pg}(\mathbf{x}^*_{1: n})$ and $\tilde{\bm{\Sigma}}_{gp}(\mathbf{x}^*_{1: n})$ are the crossing terms and while these two terms in general are not zero matrices, they are symmetric. Due to the fully occupied matrix $\tilde{\mathbf{K}}_{n_0}$, the predictive covariance with respect to the points $\tilde{\bm{\Sigma}}_p$ also contains gradient information.

\subsection{Practical Considerations}

Within this paper, we run Algorithm 1 to illustrate the performance advantage against the same settings with only vanilla GP. Nevertheless, we would like to address few remarks when this approach is applied in practice. Further studies on these considerations will be left as future work.

First, the time cost in querying derivative information should be taken into account. The methodology in obtaining derivative information could vary between problems. If we have functional evaluation of outputs and we use, for example, finite differences to estimate derivatives, there will be an additional computational cost of $O(d)$, where $d$ is the input dimension. This will be the experiment of High Pressure Fuel Supply System (section 5.2). Following derivatives computation, the complexity of the derivatives GP should also be taken into account as well. While a full GP algorithm is of order $O(n^3)$, it becomes $O((n(1+d))^3)$ with derivatives, where $n$ is number of data points. In general, the GP computation cost dominates the computational effort in our approach, as $d$ is significantly smaller than $n$. Yet, all these computational costs mentioned should also be compared to the time taken for a practical device (e.g. a flying drone) to move from one position to another. Of course, computational effort increases by incorporating derviatives, but this effort is rewareded with fewer measurements needed which is important when measurement / labelling processes are expensive. Therefore, when the measurement / labelling process is expensive and tedious, e.g. in map building and real test-bench measurements, it is worth making additional efforts for the computation while using derivatives. 

These computational costs could  be reduced accordingly via techniques such as sparse GP, as well as GPU parallelisation.

Second, when there are sufficient points where the model becomes reasonably accurate, the contribution of derivatives will begin to diminish. To this end, similar to stopping criteria for active learning (e.g. \cite{ishibashi2020stopping}), a stopping metric for using derivatives could be useful. Again, such metric very problem specific, as it depends on both the nature of the problem and derivatives estimation method. It is left as future research to figure a suitable stopping criteria on different problems.

\section{Theoretical Analysis}\label{supplementary_theory}

\subsection{Information Gain}\label{supplementary_theory_IG}
\begin{proof}[Proposition 1]
From the statement, we treat that both $IG(\mathbf{x}_{1:n})^{BALGPD}$ and $IG(\mathbf{x}_{1:n})^{BALGP}$ share the same batch of points $\mathbf{x}_{1:n}$. Despite a direct computation, few matrix properties are required to make the statement non-trivial. First, as the joint distribution is Gaussian, Equation 8 in the main text applies for both \emph{BALGP} (with matrix size $N_t$) and \emph{BALGPD} (with matrix size $N_t (1+d)$). Define $\lambda_i(\cdot)$ as the $i$th-eigenvalue of the matrix with $\lambda_1 \geq \lambda_2 \geq \cdots$ and let $\mathcal{I} := \mathbf{I}_{N_t} + \sigma^{-2} \mathbf{K}_{N_t}$ and $\tilde{\mathcal{I}} := \mathbf{I}_{N_t(1+d)} + \sigma^{-2} \tilde{\mathbf{K}}_{N_t}$ (we omit $\mathbf{x}_{1:n}$ for brevity). Then,
\begin{align*}
IG(\mathbf{x}_{1:n})^{BALGPD} - IG(\mathbf{x}_{1:n})^{BALGP}
&= \frac{1}{2} \log\det(\tilde{\mathcal{I}}) - \frac{1}{2} \log\det(\mathcal{I}) \\
&= \frac{1}{2} \left[ \log\left(\prod_{i=1}^{N_t(1+d)} \lambda_i(\tilde{\mathcal{I}}) \right) - \log \left(\prod_{i=1}^{N_t} \lambda_i(\mathcal{I}) \right) \right] \\
&= \frac{1}{2} \left[ \log\left( \frac{\prod_{i=1}^{N_t(1+d)} \lambda_i(\tilde{\mathcal{I}})}{\prod_{i=1}^{N_t} \lambda_i(\mathcal{I})} \right) \right]
\end{align*}
	
To proceed, we notice the following:
\begin{description} 
	\item[Property 1:] By construction, $\mathbf{K}_{N_t}$ is a sub-matrix of $\tilde{\mathbf{K}}_{N_t}$ when they are on the same batch of points. Then, $\mathcal{I}$ is also a sub-matrix of $\tilde{\mathcal{I}}$ because
	\begin{align*}
	\mathbf{I}_{N_t(1+d)} + \sigma^{-2} \tilde{\mathbf{K}}_{N_t} 
	&=
	\begin{pmatrix}
	\mathbf{I}_{N_t} & 0 \\
	0 & \mathbf{I}_{N_t d}
	\end{pmatrix} + \sigma^{-2} 
	\begin{pmatrix}
	\mathbf{K}_{N_t} & \left( \frac{\partial}{\partial \mathbf{x}_j} \mathbf{K}_{N_t} \right)^T \\
	\frac{\partial}{\partial \mathbf{x}_i} \mathbf{K}_{N_t} & \frac{\partial^2}{\partial \mathbf{x}_i \partial \mathbf{x}_j} \mathbf{K}_{N_t}
	\end{pmatrix} \\
	&= \begin{pmatrix}
	\mathbf{I}_{N_t} + \sigma^{-2} \mathbf{K}_{N_t} & \sigma^{-2} \left( \frac{\partial}{\partial \mathbf{x}_j} \mathbf{K}_{N_t} \right)^T \\
	\sigma^{-2} \frac{\partial}{\partial \mathbf{x}_i} \mathbf{K}_{N_t} & \mathbf{I}_{N_t d} + \sigma^{-2} \frac{\partial^2}{\partial \mathbf{x}_i \partial \mathbf{x}_j} \mathbf{K}_{N_t}
	\end{pmatrix}.
	\end{align*}
	
	\item[Property 2:] From \cite{haemers1995interlacing}, for any matrix $A$ of size $n$ and a sub-matrix $B$ of size $m$ ($m \leq n$), by \emph{Eigenvalue Interlacing Theorem}, $\lambda_i(A) \geq \lambda_i(B) \geq \lambda_{n-m+i}(A)$ for $i = 1, \cdots, m$. 
	
	\item[Property 3:] For any symmetric positive semi-definite matrix $A_n$, $\lambda_i(I + A) \geq 1$ for all $i = 1, \cdots, n$. This is because $\lambda_i(A) \geq 0$ and since such matrix is always diagonalisable. There exists a invertible matrix Q such that 
		\begin{equation*}
		QAQ^{-1} = 
		\begin{pmatrix}
		\lambda_1 & 0 & \cdots & 0 \\
		0 & \lambda_2 & \cdots & 0 \\
		\vdots & \vdots & \ddots & \vdots \\
		0 & 0 & \cdots & \lambda_n
		\end{pmatrix}
		\end{equation*}
		Then,
		\begin{equation*}
		\begin{aligned}
		Q(I+A)Q^{-1} &= QIQ^{-1} + QAQ^{-1} \\ &=
		\begin{pmatrix}
		1 & 0 & \cdots & 0 \\
		0 & 1 & \cdots & 0 \\
		\vdots & \vdots & \ddots & \vdots \\
		0 & 0 & \cdots &1
		\end{pmatrix} + 
		\begin{pmatrix}
		\lambda_1 & 0 & \cdots & 0 \\
		0 & \lambda_2 & \cdots & 0 \\
		\vdots & \vdots & \ddots & \vdots \\
		0 & 0 & \cdots & \lambda_n
		\end{pmatrix} \\ &= 
		\begin{pmatrix}
		1 + \lambda_1 & 0 & \cdots & 0 \\
		0 & 1 + \lambda_2 & \cdots & 0 \\
		\vdots & \vdots & \ddots & \vdots \\
		0 & 0 & \cdots & 1 + \lambda_n
		\end{pmatrix},
		\end{aligned}
		\end{equation*}
		which means that all eigenvalues of such matrix are no less than 1.
	\end{description}
	Then, the numerator can be decomposed into
	\begin{align*}
	\prod_{i=1}^{N_t(1+d)} \lambda_i(\tilde{\mathcal{I}}) &=
	\lambda_1(\tilde{\mathcal{I}}) \cdots \lambda_{N_t}(\tilde{\mathcal{I}}) \lambda_{N_t+1}(\tilde{\mathcal{I}}) \cdots \lambda_{N_t(1+d)}(\tilde{\mathcal{I}}) \\
	&\geq \lambda_1(\mathcal{I}) \cdots \lambda_{N_t}(\mathcal{I}) \lambda_{N_t+1}(\tilde{\mathcal{I}}) \cdots \lambda_{N_t(1+d)}(\tilde{\mathcal{I}}) \\
	&= \left( \prod_{i=1}^{N_t} \lambda_i(\mathcal{I})  \right) \lambda_{N_t+1}(\tilde{\mathcal{I}}) \cdots \lambda_{N_t(1+d)}(\tilde{\mathcal{I}}).
	\end{align*}
	Hence, with Property 3 that all remaining terms are no less than 1, the difference is no less than zero and we have our claim.
\end{proof}

\begin{remark}
It is possible to argue the proof based on the fact that the mutual information is a monotonic set function. Nevertheless, the advantage of direction computation is that we discover the exact magnitude of the difference between the two mutual informations. Our computation also gives the lower bound of the difference and they are governed by the product of extended eigenvalues.
\end{remark}

\begin{proof}[Proposition 2]
	In AL regression, our regime is greedy such that we choose points in order that the optimality criterion is maximised. As observations are independent given functional evaluations, the information gain satisfies a diminishing return property called \emph{submodularity} (it is indeed \emph{monotone submodular}, see \cite{krause2014submodular}). Then, by \cite{nemhauser1978analysis}, 
	\begin{equation}\label{greedy_max}
	I(\lbrace y^*_{1: n, \tau} \rbrace^t_{\tau=1} ;\lbrace \mathbf{f}_{1: n, \tau} \rbrace^t_{\tau=1}) \geq (1 - \exp(-1)) \tilde{\gamma}_t.
	\end{equation}	
	Since we already have the explicit form of the information gain, the remaining steps are to find an upper bound of the expression in Equation 8 in the main text. The idea is similar to Lemma 1 in Seeger et al.\cite{seeger2008information} and Lemma 7.6 in Srinivas et al.\cite{srinivas2012information}. First, by Mercer's Theorem we decompose the covariance matrix $\tilde{\mathbf{K}}_{N_t}$ as
	\begin{equation}
	\tilde{\mathbf{K}}_{N_t} = \bar{\bm{\Phi}}_S \bm{\Lambda}_S \bar{\bm{\Phi}}^T_S.
	\end{equation}
	On contrary to Lemma 1 in Seeger et al.\cite{seeger2008information}, the key difference is that the eigenfunctions $\bar{\bm{\Phi}}_S$ include derivatives information. This matrix is of size $N_t(1+d) \times S$ (for $S$ finite, infinite $S$ can be treated similarly but we omit here for simplicity) and explicitly, 
	\begin{equation*}
	\bar{\bm{\Phi}}_S = 
	\begin{pmatrix}
	\begin{pmatrix}
	\mathds{1} \\ 
	\frac{\partial}{\partial x_{i, 1}} \\
	\vdots \\
	\frac{\partial}{\partial x_{i, d}}
	\end{pmatrix} \phi_s(\mathbf{x}_i)
	\end{pmatrix}^{N_t, S}_{i = 1, s = 1}
	= 
	\begin{pmatrix}
	\phi_1(\mathbf{x}_1) & \cdots & \phi_s(\mathbf{x}_1) \\
	\frac{\partial}{\partial x_{1, 1}} \phi_1(\mathbf{x}_1) & \cdots & \frac{\partial}{\partial x_{1, 1}} \phi_s(\mathbf{x}_1) \\
	\vdots & & \vdots \\
	\frac{\partial}{\partial x_{1, d}} \phi_1(\mathbf{x}_1) & \cdots & \frac{\partial}{\partial x_{1, d}} \phi_s(\mathbf{x}_1) \\
	\vdots & & \vdots \\
	\phi_1(\mathbf{x}_N) & \cdots & \phi_s(\mathbf{x}_{N_t}) \\
	\frac{\partial}{\partial x_{N_t, 1}} \phi_1(\mathbf{x}_{N_t}) & \cdots & \frac{\partial}{\partial x_{N_t, 1}} \phi_s(\mathbf{x}_{N_t}) \\
	\vdots & & \vdots \\
	\frac{\partial}{\partial x_{N_t, d}} \phi_1(\mathbf{x}_{N_t}) & \cdots & \frac{\partial}{\partial x_{N_t, d}} \phi_s(\mathbf{x}_{N_t}) \\
	\end{pmatrix}.
	\end{equation*}
	
	Then, by multiplying with the diagonal matrices, the following $S \times S$ matrix is given by
	\begin{equation*}
	\bm{\Lambda}^{\frac{1}{2}}_S \bar{\bm{\Phi}}^T_S \bar{\bm{\Phi}}_S \bm{\Lambda}^{\frac{1}{2}}_S = 
	\begin{pmatrix}
	\sqrt{\lambda_s \lambda_t} \sum^{N_t}_{i = 1} \left( \phi_s(\mathbf{x}_i) \phi_t(\mathbf{x}_i) + \sum^d_{j = 1} \frac{\partial}{\partial x_{i, j}} \phi_s(\mathbf{x}_i) \frac{\partial}{\partial x_{i, j}} \phi_t(\mathbf{x}_i) \right)
	\end{pmatrix}^S_{s, t = 1}.
	\end{equation*}
	
	We can now take the bound of the information gain using Weinstein–Aronszajn identity and Hadamard's inequality (Theorem 7.8.1 in \cite{horn2012matrix}):
	\begin{align*}
	\log \det \left( \mathbf{I}_{N_t(1+d)} + \sigma^{-2} \tilde{\mathbf{K}}_{N_t} \right) 
	&= \log \det \left( \mathbf{I}_{N_t(1+d)} + \sigma^{-2} \bar{\bm{\Phi}}_S \bm{\Lambda}_S \bar{\bm{\Phi}}^T_S \right) \\
	&= \log \det \left( \mathbf{I}_{S} + \sigma^{-2} \bm{\Lambda}^{\frac{1}{2}}_S \bar{\bm{\Phi}}^T_S  \bar{\bm{\Phi}}_S \bm{\Lambda}^{\frac{1}{2}}_S \right) \quad  \mbox{(Weinstein–Aronszajn identity)} \\
	&\leq \log \det \left( \diag\left( \mathbf{I}_S + \sigma^{-2} \bm{\Lambda}^{\frac{1}{2}}_S \bar{\bm{\Phi}}^T_S \bar{\bm{\Phi}}_S \bm{\Lambda}^{\frac{1}{2}}_S \right) \right) \quad \mbox{(Hadamard's inequality)} \\
	&= \log \left( \prod_{s \geq 1} \left( 1 + \sigma^{-2} \lambda_s \sum^{N_t}_{i = 1} \left( \left( \phi_s(\mathbf{x}_i) \right)^2 + \sum^d_{j = 1} \left( \frac{\partial}{\partial x_{i, j}} \phi_s(\mathbf{x}_i) \right)^2 \right) \right) \right) \\
	&= \sum_{s \geq 1} \log \left( 1 + \sigma^{-2} \lambda_s \sum^{N_t}_{i = 1} \left( \left( \phi_s(\mathbf{x}_i) \right)^2 + \sum^d_{j = 1} \left( \frac{\partial}{\partial x_{i, j}} \phi_s(\mathbf{x}_i) \right)^2 \right) \right).
	\end{align*}
	With the diminishing return property, we finally have
	\[  \tilde{\gamma}_t \leq \frac{1}{2 (1 - \exp(-1))} \sum_{s \geq 1} \log \left( 1 + \sigma^{-2} \lambda_s \sum^{N_t}_{i = 1} \left( \left( \phi_s(\mathbf{x}_i) \right)^2 + \sum^d_{j = 1} \left( \frac{\partial}{\partial x_{i, j}} \phi_s(\mathbf{x}_i) \right)^2 \right) \right). \]
\end{proof}

\subsection{Decay in Predictive Covariance Matrix}\label{appendix_DetailsTheory_part2}
In this subsection, we first present the concept of \emph{L\"{o}wner partial ordering} with the relevant proposition before proofing our propositions.
\begin{definition}[L\"{o}wner partial ordering (\cite{horn2012matrix}, section 7.7)] \label{def_Louewner}
	For any two $n \times n$ matrices $\mathbf{K}_1$ and $\mathbf{K}_2$, we write $\mathbf{K}_1 \succeq \mathbf{K_2}$ (or equivalently, $\mathbf{K}_2 \preceq \mathbf{K}_1$) if both $\mathbf{K}_1$ and $\mathbf{K}_2$ are \emph{symmetric} and $\mathbf{K}_1 - \mathbf{K}_2$ is \emph{positive semidefinite} (or equivalently,  $\mathbf{K}_2 - \mathbf{K}_1$ is \emph{negative semidefinite}). A similar notion can be defined for positive definite (and negative definite) matrix: $\mathbf{K}_1 \succ \mathbf{K_2}$ if both $\mathbf{K}_1$ and $\mathbf{K}_2$ are \emph{symmetric} and $\mathbf{K}_1 - \mathbf{K}_2$ is \emph{positive definite}. 
\end{definition}
\begin{remark}
	A matrix $\mathbf{K}$ which is symmetric and positive semidefinite can be written as $\mathbf{K} \succeq 0$ (or $\mathbf{K} \succ 0$ if $\mathbf{K}$ is positive definite). Negative definite matrix and negative semidefinite matrix are defined as $\mathbf{K} \preceq 0$ and $\mathbf{K} \prec 0$)
\end{remark}
We can treat this notation as an extension to ordering between real numbers. In particular, this ordering preserves after taking determinant, eigenvalue and trace from a matrix to a real number. The ordering is reversed after taking the inverse of both matrices, as both matrices are invertible. This can be formalised in the following proposition:

\begin{proposition} \label{prop_LPO}
	For any two $n \times n$ matrices $\mathbf{K}_1$ and $\mathbf{K}_2$ which are symmetric. Let $\lambda_1(\mathbf{K}_1) \geq \cdots \geq \lambda_n(\mathbf{K}_1)$ and $\lambda_1(\mathbf{K}_2) \geq \cdots \geq \lambda_n(\mathbf{K}_2)$ are the order eigenvalues of $\mathbf{K}_1$ and $\mathbf{K}_2$ respectively. Then,
	\begin{enumerate}
		\item If $\mathbf{K}_1 \succ 0$ and $\mathbf{K}_2 \succ 0$, then $\mathbf{K}_1 \succeq \mathbf{K}_2$ if and only if $\mathbf{K}^{-1}_1 \preceq \mathbf{K}^{-1}_2$.
		\item If $\mathbf{K}_1 \succeq \mathbf{K}_2$, then $\lambda_i(\mathbf{K}_1) \geq \lambda_i(\mathbf{K}_2)$ for $i = 1, \cdots, n$.
		\item If $\mathbf{K}_1 \succeq \mathbf{K}_2$, then $\tr(\mathbf{K}_1) \geq \tr(\mathbf{K}_2)$ with equality if and only if $\mathbf{K}_1 = \mathbf{K}_2$.
		\item If $\mathbf{K}_1 \succeq \mathbf{K}_2 \succeq 0$, then $\det(\mathbf{K}_1) \geq \det(\mathbf{K}_2) \geq 0$.
	\end{enumerate}
\end{proposition}
\begin{proof}
	The proof for this proposition is given in Horn et al. (\cite{horn2012matrix}, corollary 7.7.4).
\end{proof}

\begin{proof}[Theorem 1]
	This is an application from the variational characterisation of Schur complement (\cite{lami2016schur}, theorem 1). For $\mathbf{x}_{1:n_0}, \mathbf{x}_{1:n} \in \mathcal{X}$ (recall that $\mathbf{x}_{1:n_0}$ are initial points and $\mathbf{x}_{1:n}$ are new batch of explored points), the full covariance matrix of the joint distribution with derivative information is expressed by, 
	\begin{equation}
	\tilde{\mathcal{K}} := 
	\begin{pmatrix}
	\mathbf{K}_{n_0} + \sigma^2 \mathbf{I}_{n_0} & \left( \frac{\partial}{\partial \mathbf{x}_j} \mathbf{K}_{n_0} \right)^T & k(\mathbf{x}_{1: n_0}, \mathbf{x}_{1:n})  \\
	\frac{\partial}{\partial \mathbf{x}_i} \mathbf{K}_{n_0} & \frac{\partial^2}{\partial \mathbf{x}_i \partial \mathbf{x}_j} \left( \mathbf{K}_{n_0} \right) + \sigma^2 \mathbf{I}_{n_0d} & \frac{\partial}{\partial \mathbf{x}_i} k(\mathbf{x}_{1: n_0}, \mathbf{x}_{1:n}) \\
	k(\mathbf{x}_{1: n_0}, \mathbf{x}_{1:n})^T & \left( \frac{\partial}{\partial \mathbf{x}_i} k(\mathbf{x}_{1: n_0}, \mathbf{x}_{1:n}) \right)^T & k(\mathbf{x}_{1:n}. \mathbf{x}_{1:n})
	\end{pmatrix},
	\end{equation}
	
	Also, the corresponding matrix without derivative information is given by
	\begin{equation}
	\mathcal{K} :=
	\begin{pmatrix}
	\mathbf{K}_{n_0} + \sigma^2 \mathbf{I}_{n_0} & k(\mathbf{x}_{1: n_0}, \mathbf{x}_{1:n})  \\
	k(\mathbf{x}_{1: n_0}, \mathbf{x}_{1:n})^T & k(\mathbf{x}_{1:n}, \mathbf{x}_{1:n})
	\end{pmatrix}
	\end{equation}
	
	We recall that by construction, both matrices $\mathcal{K}$ and $\tilde{\mathcal{K}}$ are positive semi-definite and symmetric. Then, motivated by 7.7.P40 of Horn et al. \cite{horn2012matrix}, the predictive covariance matrix $\bm{\Sigma}(\mathbf{x}_{1:n})$ and $\tilde{\bm{\Sigma}}_p(\mathbf{x}_{1:n})$ are precisely the Schur complement of the two matrices and they can be expressed as 
	\begin{align}
	\bm{\Sigma}(\mathbf{x}_{1:n}) &= \max\lbrace E \in M_{n \times n}, E = E^T \mbox{ and } \mathcal{K} \succeq \mathbf{0}_{n _0 \times n_0} \oplus E \rbrace, \\
	\tilde{\bm{\Sigma}}_p(\mathbf{x}_{1:n}) &= \max\lbrace E \in M_{n \times n}, E = E^T \mbox{ and } \tilde{\mathcal{K}} \succeq \mathbf{0}_{n_0(1+d) \times n_0(1+d)} \oplus E \rbrace,
	\end{align}
	where $\oplus$ is the direct sum of two matrices (for two matrices $A$ (size $m \times n$) and $B$ (size $p \times q$), the direct sum $A \oplus B$ is a $(m + n) \times (p + q)$ matrix given as $\begin{pmatrix}
	A & \mathbf{0} \\ \mathbf{0} & B
	\end{pmatrix}$) and the maximum of these sets are with respect to the L\"{o}wner partial ordering. The set $\lbrace E \in M_{n \times n}, E = E^T \mbox{ and } \tilde{\mathcal{K}} \succeq \mathbf{0}_{n_0(1+d) \times n_0(1+d)} \oplus E \rbrace \subseteq \lbrace E \in M_{n \times n}, E = E^T \mbox{ and } \mathcal{K} \succeq \mathbf{0}_{n_0 \times n_0} \oplus E \rbrace$, because for any matrix $E$ satisfying $\tilde{\mathcal{K}} \succeq \mathbf{0}_{n_0(1+d) \times n_0(1+d)} \oplus E$, given a vector $\mathbf{v}_{1, n_0}$ of size $n_0$, zero vector $\mathbf{0}_{n_0d}$ of size $n_0d$ and another vector $\mathbf{v}_{2, n}$ of size $n$,
	\begin{equation*}
	\begin{pmatrix}
	\mathbf{K}_{n_0} + \sigma^2 \mathbf{I}_{n_0} & \left( \frac{\partial}{\partial \mathbf{x}_j} \mathbf{K}_{n_0} \right)^T & k(\mathbf{x}_{1: n_0}, \mathbf{x}_{1:n})  \\
	\frac{\partial}{\partial \mathbf{x}_i} \mathbf{K}_{n_0} & \frac{\partial^2}{\partial \mathbf{x}_i \partial \mathbf{x}_j} \left( \mathbf{K}_{n_0} \right) + \sigma^2 \mathbf{I}_{n_0d} & \frac{\partial}{\partial \mathbf{x}_i} k(\mathbf{x}_{1: n_0}, \mathbf{x}_{1:n}) \\
	k(\mathbf{x}_{1: n_0}, \mathbf{x}_{1:n})^T & \left( \frac{\partial}{\partial \mathbf{x}_i} k(\mathbf{x}_{1: n_0}, \mathbf{x}_{1:n}) \right)^T & k(\mathbf{x}_{1:n}. \mathbf{x}_{1:n}) - E
	\end{pmatrix} \succeq \mathbf{0}
	\end{equation*}
	\begin{equation*}
	\resizebox{1.0 \textwidth}{!}{$
		\Leftrightarrow  
		\begin{pmatrix}
		\mathbf{v}^T_{1, n_0} & \mathbf{0}^T_{n_0d} & \mathbf{v}^T_{2, n}
		\end{pmatrix}
		\begin{pmatrix}
		\mathbf{K}_{n_0} + \sigma^2 \mathbf{I}_{n_0} & \left( \frac{\partial}{\partial \mathbf{x}_j} \mathbf{K}_{n_0} \right)^T & k(\mathbf{x}_{1: n_0}, \mathbf{x}_{1:n})  \\
		\frac{\partial}{\partial \mathbf{x}_i} \mathbf{K}_{n_0} & \frac{\partial^2}{\partial \mathbf{x}_i \partial \mathbf{x}_j} \left( \mathbf{K}_{n_0} \right) + \sigma^2 \mathbf{I}_{n_0d} & \frac{\partial}{\partial \mathbf{x}_i} k(\mathbf{x}_{1: n_0}, \mathbf{x}_{1:n}) \\
		k(\mathbf{x}_{1: n_0}, \mathbf{x}_{1:n})^T & \left( \frac{\partial}{\partial \mathbf{x}_i} k(\mathbf{x}_{1: n}, \mathbf{x}_{1:n}) \right)^T & k(\mathbf{x}_{1:n}. \mathbf{x}_{1:n}) - E
		\end{pmatrix}
		\begin{pmatrix}
		\mathbf{v}_{1, n_0} \\ \mathbf{0}_{n_0d} \\ \mathbf{v}_{2, n}
		\end{pmatrix} \succeq 0 $}
	\end{equation*}
	\begin{equation*}
	\Leftrightarrow \begin{pmatrix}
	\mathbf{v}^T_{1, n_0} & \mathbf{v}^T_{2, n}
	\end{pmatrix}
	\begin{pmatrix}
	\mathbf{K}_{n_0} + \sigma^2 \mathbf{I}_{n_0} & k(\mathbf{x}_{1: n_0}, \mathbf{x}_{1:n})  \\
	k(\mathbf{x}_{1: n_0}, \mathbf{x}_{1:n})^T & k(\mathbf{x}_{1:n}. \mathbf{x}_{1:n}) - E
	\end{pmatrix}
	\begin{pmatrix}
	\mathbf{v}_{1, n_0} \\ \mathbf{v}_{2, n}
	\end{pmatrix} \succeq 0
	\end{equation*}
	
	Therefore, the maximum value of a larger set is always no smaller than the smaller set and this holds under the L\"{o}wner partial ordering. That is,
	\begin{align*}
	\bm{\Sigma}(\mathbf{x}_{1:n}) &= \max\lbrace E \in M_{n \times n}, E = E^T \mbox{ and } \mathcal{K} \succeq \mathbf{0}_{n_0 \times n_0} \oplus E \rbrace \\
	&\succeq \max\lbrace E \in M_{n \times n}, E = E^T \mbox{ and } \tilde{\mathcal{K}} \succeq \mathbf{0}_{n_0(1+d) \times n_0(1+d)} \oplus E \rbrace \\
	&= \tilde{\bm{\Sigma}}_p(\mathbf{x}_{1:n}).
	\end{align*}
\end{proof}

For theorem 2, since the case of \emph{determinant} follows from the combination of the previous theorem and theorem 2 of \cite{zimmer2018safe} and for the case of \emph{maximum eigenvalue}, it follows from the fact that the trace is the sum of all eigenvalues, our focus here is for the case of \emph{trace}. The plan of attack is as follow: We first provide an upper bound for the average of the trace of covariance matrix without derivatives, then we use Theorem 1 to connect the corresponding matrix with derivative information. To proceed, we need the following lemma:

\begin{lemma} \label{prop2_sec42_lemma1}
	For any arbitrary sequence $\mathbf{x}_{1: n, \tau}$, after observing for $t$ rounds, the average of the trace of the covariance matrix is upper bounded by 
	\begin{equation} \label{prop2_sec42_lemma1_eq}
	\frac{1}{t} \sum^t_{\tau=1} \tr\left( \mathbf{\Sigma}_{\tau-1} (\mathbf{x}_{1: n, \tau}) \right) \leq C_{\mbox{tr}} \frac{\gamma_t}{t},
	\end{equation}
	where $C_{\mbox{tr}} = 2\sigma^2 (1 + n \sigma^{-2} \sigma^2_f)$ is a constant and $\mathbf{\Sigma}_{\tau-1} (\mathbf{x}_{1: n, \tau})$ the $\tau$-th predictive covariance trained on it.
\end{lemma}
\begin{proof}
	First, we denote the eigenvalues of any positive semidefinite matrix $A$ of size $n$ as $\lambda_1(A) \geq \cdots \geq \lambda_n(A) \geq 0$. Then, since the geometric mean is always no more than the arithmetic mean, the maximal eigenvalue and the trace of the predictive covariance matrix are related as follow
	\begin{equation*}
	\lambda_1\left( \mathbf{\Sigma}_{t-1}(\mathbf{x}_{1: n, \tau}) \right) \leq \tr \left( \mathbf{\Sigma}_{t-1}(\mathbf{x}_{1: n, \tau}) \right) \leq n \sigma^2_f.
	\end{equation*}
	
	We now look into the term $\log\det\left( \mathbf{I}_{n} + \sigma^{-2} \mathbf{\Sigma}_{\tau-1}(\mathbf{x}_{1:n, \tau}) \right)$ and we can lower bound this expression:
	
	\begin{align*}
	&\quad \log\det\left( \mathbf{I}_{n} + \sigma^{-2} \mathbf{\Sigma}_{\tau-1}(\mathbf{x}_{1: n, \tau}) \right) \\
	&= \log\left( \prod^n_{j = 1} \lambda_j\left( \mathbf{I}_{n} + \sigma^{-2} \mathbf{\Sigma}_{\tau-1}(\mathbf{x}_{1:n, \tau}) \right) \right) \\
	&= \sum^n_{j = 1} \log \left( \lambda_j \left( \mathbf{I}_{n} + \sigma^{-2} \mathbf{\Sigma}_{\tau-1}(\mathbf{x}_{1: n, \tau}) \right) \right) \\
	&\geq \sum^n_{j = 1} \frac{\lambda_j\left( \mathbf{I}_{n} + \sigma^{-2} \mathbf{\Sigma}_{\tau-1}(\mathbf{x}_{1:n, \tau}) \right) - 1}{\lambda_j\left( \mathbf{I}_{n} + \sigma^{-2} \mathbf{\Sigma}_{\tau-1}(\mathbf{x}_{1:n, \tau}) \right)} \\
	&\geq \frac{1}{\lambda_1\left( \mathbf{I}_{n} + \sigma^{-2} \mathbf{\Sigma}_{\tau-1}(\mathbf{x}_{1:n, \tau}) \right)} \sum^n_{j = 1} \left( \lambda_j\left( \mathbf{I}_{n} + \sigma^{-2} \mathbf{\Sigma}_{\tau-1}(\mathbf{x}_{1: n, \tau}) \right) - 1\right) \\
	&= \frac{\sigma^{-2}}{\lambda_1\left( \mathbf{I}_{n} + \sigma^{-2} \mathbf{\Sigma}_{\tau-1}(\mathbf{x}_{1:n, \tau}) \right)} \tr(\mathbf{\Sigma}_{\tau-1}(\mathbf{x}_{1:n, \tau}))) \\
	&\geq \frac{\sigma^{-2}}{\lambda_1(\mathbf{I}_n) + \lambda_1(\sigma^{-2} \mathbf{\Sigma}_{\tau-1}(\mathbf{x}_{1:n, \tau}))} \tr(\mathbf{\Sigma}_{t-1}(\mathbf{x}_{1:n, \tau})) \quad (\mbox{Weyl's inequality}) \\
	&= \frac{\sigma^{-2}}{1 + \sigma^{-2} \lambda_1(\tilde{\Sigma}_{\tau-1}(\mathbf{x}_{1:n, \tau}))} \tr(\mathbf{\Sigma}_{\tau-1}(\mathbf{x}_{1:n, \tau})) \\
	&\geq \frac{\sigma^{-2}}{1 + \sigma^{-2} n \sigma^2_f} \tr(\mathbf{\Sigma}_{\tau-1}(\mathbf{x}_{1:n, \tau})) 
	\end{align*}
	
	Rearrange the above inequality we have
	\begin{equation*}
	\tr(\mathbf{\Sigma}_{\tau-1}(\mathbf{x}_{1:n, \tau})) \leq \sigma^2(1 + n\sigma^{-2}\sigma^2_f)\log\det\left( \mathbf{I}_n + \sigma^{-2} \mathbf{\Sigma}_{\tau-1}(\mathbf{x}_{1:n, \tau}) \right).
	\end{equation*}
	
	Using the definition of $\tilde{\gamma}_t$ under GP distribution (\cite{srinivas2012information, zimmer2018safe}) and Lemma \ref{prop2_sec42_lemma1}:  
	\begin{align*}
	\frac{1}{t} \sum^t_{\tau=1} \tr(\mathbf{\Sigma}_{\tau-1}(\mathbf{x}_{1: n, \tau}))
	&\leq \frac{1}{t} \sum^t_{\tau=1} \sigma^2(1 + n\sigma^{-2}\sigma^2_f)\log\det\left( \mathbf{I}_n + \sigma^{-2} \mathbf{\Sigma}_{\tau-1}(\mathbf{x}_{1:n, \tau}) \right) \\
	&= \frac{2 \sigma^2(1 + n\sigma^{-2}\sigma^2_f)}{t} \cdot \frac{1}{2} \sum^t_{\tau=1} \log\det\left( \mathbf{I}_n + \sigma^{-2} \mathbf{\Sigma}_{\tau-1}(\mathbf{x}_{1:n, \tau}) \right) \\
	&= \frac{2 \sigma^2(1 + n\sigma^{-2}\sigma^2_f)}{t} I(\lbrace y^*_{1:n, \tau} \rbrace^t_{\tau=1} ;\lbrace \mathbf{f}_{1:n, \tau} \rbrace^t_{\tau=1}) \\
	&\leq C_{\tr} \frac{\gamma_t}{t}.
	\end{align*}
	where $C_{\tr} = 2 \sigma^2(1 + n\sigma^{-2}\sigma^2_f)$ is the constant we desire.
\end{proof}

\begin{proof}[Theorem 2]
	With lemma 1 we can proceed to the actual proof. As it does not specify whether the sequence comes with or without derivatives information, the statement actually holds for any given set of points. Therefore, we can also write
	\begin{equation}
	\frac{1}{t} \sum^t_{\tau=1} \tr(\tilde{\mathbf{\Sigma}}_{p, \tau-1}(\tilde{\mathbf{x}}_{1:n, \tau})) \leq \frac{1}{t} \sum^t_{\tau=1} \tr(\tilde{\mathbf{\Sigma}}_{p, \tau-1} (\mathbf{x}^*_{1:n, \tau})).
	\end{equation}
From Theorem 1 and properties of L\"{o}wner partial ordering, 
	\begin{equation}
	\tr(\tilde{\mathbf{\Sigma}}_{p, \tau-1}(\mathbf{x}^*_{1:n, \tau})) \leq \tr(\mathbf{\Sigma}_{\tau-1}(\mathbf{x}^*_{1:n, \tau}))
	\end{equation}
and as lemma 1 holds for any sequence, we finally use Theorem 5 in Srinivas et al.\cite{srinivas2012information}:
	\begin{equation}
	\frac{1}{t} \sum^t_{\tau=1} \tr(\tilde{\mathbf{\Sigma}}_{p, \tau-1}(\tilde{\mathbf{x}}_{1:n, \tau})) = \mathcal{O}\left( \frac{(\log(t))^{d+1}}{t} \right).
	\end{equation}
\end{proof}

\setcounter{figure}{7}
\section{Empirical Evaluation}\label{supplementary_eval}

\subsection{Simulated Function} 
\label{supplementary_eval_exp1}
The cardinal sine function from Schreiter et al.\cite{schreiter2015safe}, as well as its derivative, are given by
\begin{equation*}
f(x) = \frac{10 \sin(x-10)}{x-10}, \qquad f'(x) = \frac{10 \cos(x-10)}{x-10} - \frac{10 \sin(x-10)}{(x-10^2)}.
\end{equation*}

\subsubsection*{Further diagrams}

\begin{figure*}[h]
	\centering
	\includegraphics[scale=.22]{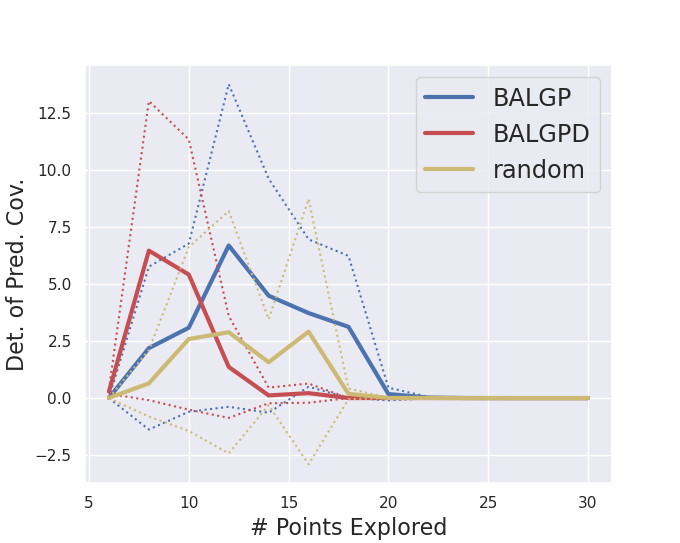}
	\includegraphics[scale=.22]{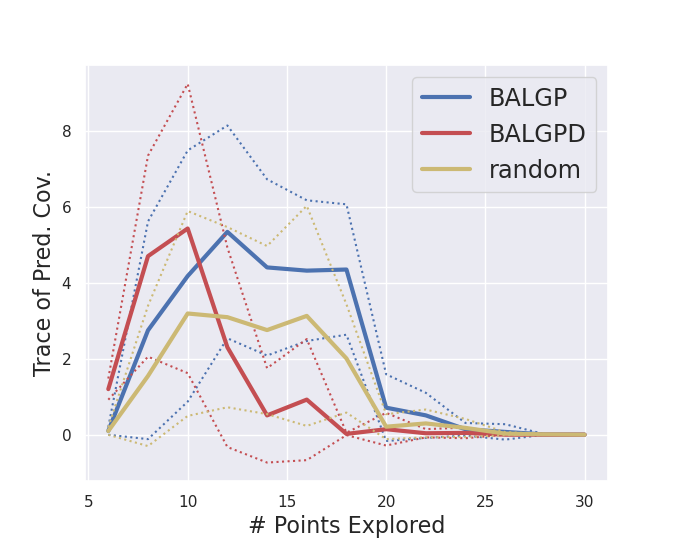}
	\includegraphics[scale=.22]{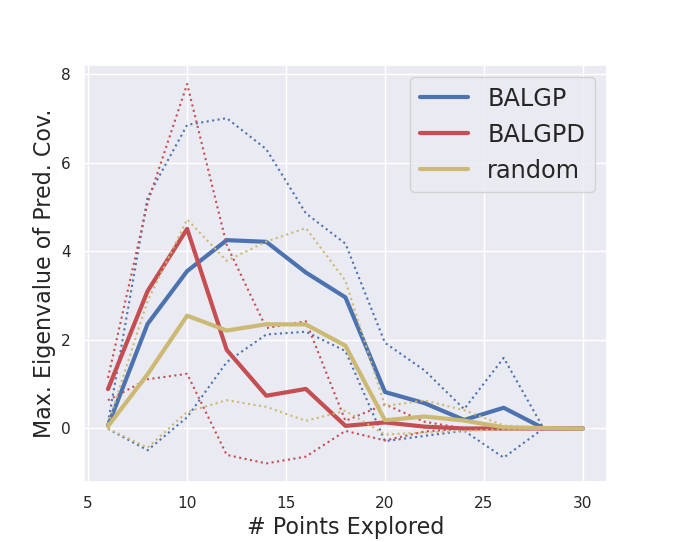}
	\caption{\emph{Simulated function}: The three diagrams plot the determinant (left), trace (middle) and maximum eigenvalue (right) attained under random (yellow), \emph{BALGPD} (red) and \emph{BALGP} (blue) as more points are explored.}
	\label{diagram_exp1_det_tr_maxeig_extra}
\end{figure*}

We also present the behaviour of the trace, maximum eigenvalue and determinant of the predictive covariance attained under random regime, \emph{BALGP} and \emph{BALGPD} in figure \ref{diagram_exp1_det_tr_maxeig_extra}. In practice, although the actual points explored are different under different strategies, we observe that under the three optimality criterion, the value attained under \emph{BALGPD} is eventually lower than that under \emph{BALGP}, which shows that theorem 1 is still by and large preserved.

\subsection{High Pressure Fluid System} 
\label{supplementary_eval_exp2}

\subsubsection*{BALGPD plus Safety}
\begin{algorithm}[h]
	\caption{Batch Active Leaning using Gaussian Process (Regression) with derivatives, plus safety constraint.}\label{alGPDSafety}
	\begin{algorithmic}
		\STATE {\bfseries Input:} Initial training data $\{\mathbf{x}_{1:n_0}, y_{1: n_0}, \nabla y_{1:n_0}\}$, batch size $n_t$ for each round. Plus an initial safety model with corresponding initial safety values $z_{1: n_0}$.
		\STATE Train the initial exploration GP model with derivatives, as well as the safety model.
		\FOR{round $t$ = 1 to $T$}
		\STATE {1.} Sample initial $n_t$ points: $\mathbf{x}_{1: n_t}$.
		\STATE {2.} Solve equation 14 in the main text for the optimised new batch of points $\mathbf{x}^*_{1: n_t}$. The predictive covariance matrix is evaluated based on the GP derivative model as in equation 4, 5 and 6 in the main text.
		\STATE {3.} At inputs $\mathbf{x}^*_{1: n_t}$, evaluate $y^*_{1: n_t}$ and $\nabla y^*_{1: n_t}$. Also, evaluate the corresponding safety values $z_{1: n_t}$ from the new batch of points via the physical system. 
		\STATE {4.} Update the training data set and the safety values data. 
		\STATE {5.} Update GP with derivatives regression model in equation 3 in the main text, based on the initial information, explored patch of points and the corresponding evaluations up to round $t$. The safety model should also be updated with the new safety values included.
		\ENDFOR
		\STATE {\bfseries Output:} Set of predictive outputs with its gradient and the GP with derivative model describing the function. Safe region can also be recorded.
	\end{algorithmic}
\end{algorithm}

\begin{remark}
In this experiment, we run \emph{BALGPD} for exploration of outputs and \emph{BALGP} for safety exploration. It is possible to 1) run \emph{BALGPD} on the safety exploration as well, or 2) Use other evaluation regimes instead of GP based model.
\end{remark}

\subsubsection*{Further Diagrams}

\begin{figure*}[h]
	\centering
	\includegraphics[scale=.22]{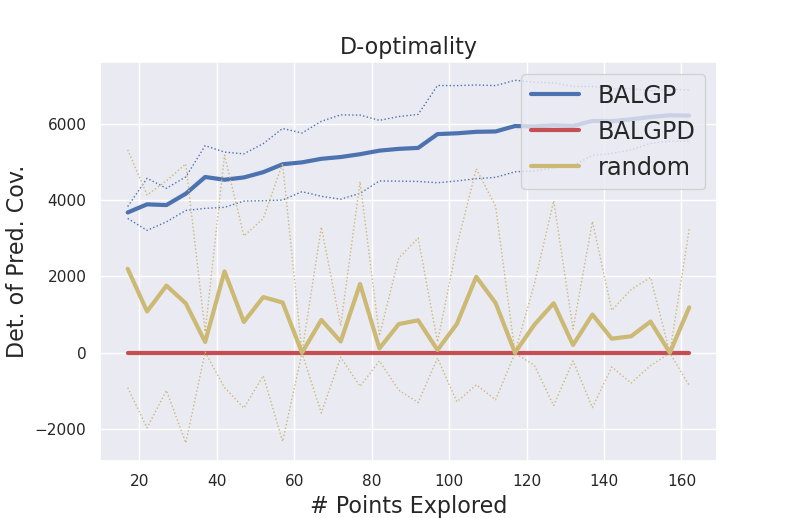}
	\includegraphics[scale=.22]{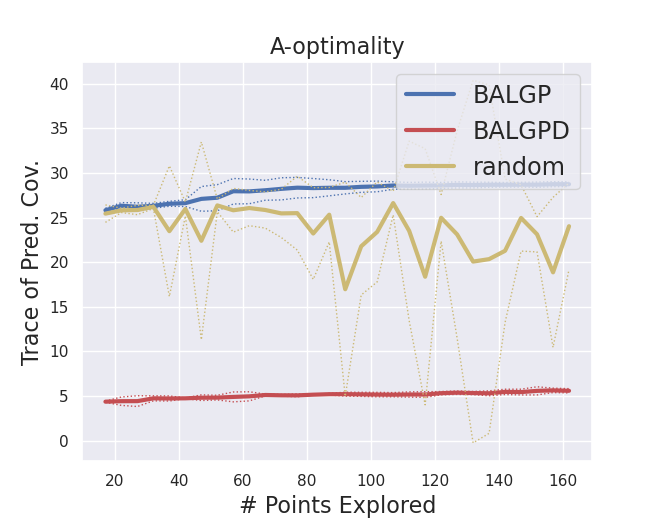}
	\includegraphics[scale=.22]{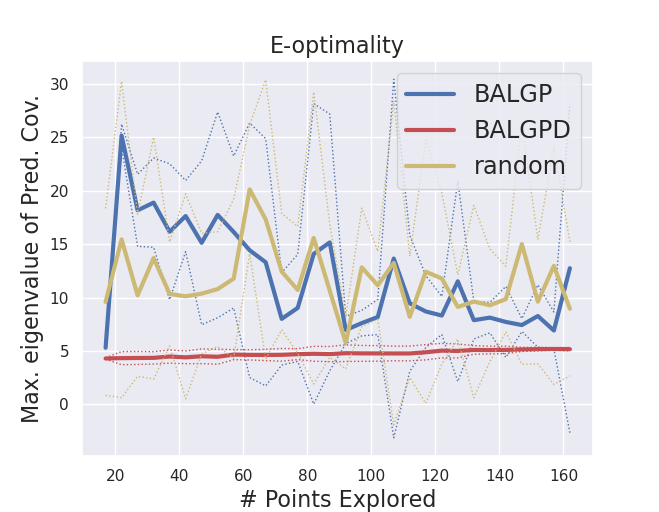}
	\caption{\emph{High Pressure Fuel Supply System}: The three diagrams plot the determinant (left), trace (middle) and maximum eigenvalue (right) attained under \emph{BALGPD} (red), \emph{BALGP} (blue) and randomised exploration (yellow) as more points are explored.}
	\label{diagram_exp2_trANDmaxeig_extra}
\end{figure*}

Figure \ref{diagram_exp2_trANDmaxeig_extra} presents how the trace and maximum eigenvalue of the GP predictive covariance change as more points are explored. Note that $k(\cdot, \cdot) > 1$ at some positions, which violate the assumption in Theorem 2 in the main text. Nevertheless, not only does \emph{BALGPD} attains a better performance than \emph{BALGP} (as shown in the main text), \emph{BALGPD} also attains a smaller predictive covariance under all three optimality criterion than that under \emph{BALGP}. This shows our algorithm is persistent in predictive performance even when some theoretical assumption do not meet.

\subsection{Map Reconstruction} 
\label{supplementary_eval_exp3}

\subsubsection*{Gradient Estimation}

Based on Meyer et al.\cite{meyer2001gradient}, we select 4 closest neighbouring points to estimate the gradient. Scattered data is generally the case when doing landscape measurements. In order to estimate the gradient, a set of distinct sample points (with at least one point being non-collinear) are selected forming a neighbourhood around the point of interest and 4 closest neighbouring points are selected. Let this point be $\mathbf{p} = (x_1, x_2, y)$. Suppose there are in total $n$ points to select around $\mathbf{p}$, we look into $\mathbf{p}_1, \cdots, \mathbf{p}_{n}$, compute $\mathbf{v}_i = \mathbf{p}_i - \mathbf{p}$ for $i = 1, \cdots, n$ such that the 4 points with the smallest distances are selected. We roughly approximate the partial derivatives as 
\[ y_i - y = \frac{\partial y}{\partial x_1} (x_{i, 1} - x_1) + \frac{\partial y}{\partial x_2} (x_{i, 2} - x_2). \]
Therefore, the quantity we are interested are $\frac{\partial y}{\partial x_1}$ and $\frac{\partial y}{\partial x_2}$. The equation we would like to solve is $\sigma = \mathbf{V} \Delta$, defined as
\begin{equation*}
\begin{pmatrix}
\frac{y_1 - y}{\| \mathbf{v}_1 \|} \\
\frac{y_2 - y}{\| \mathbf{v}_2 \|} \\
\frac{y_3 - y}{\| \mathbf{v}_3 \|} \\
\frac{y_4 - y}{\| \mathbf{v}_4 \|}
\end{pmatrix} = 
\begin{pmatrix}
x_{1, 1} - x_1 & x_{1, 2} - x_2 \\
x_{2, 1} - x_1 & x_{2, 2} - x_2 \\
x_{3, 1} - x_1 & x_{3, 2} - x_2 \\
x_{4, 1} - x_1 & x_{4, 2} - x_2
\end{pmatrix} 
\begin{pmatrix}
\frac{\partial y}{\partial x_1} \\
\frac{\partial y}{\partial x_2} 
\end{pmatrix}.
\end{equation*}

By rearranging the equation, $\Delta = (\mathbf{V}^T \mathbf{V})^{-1} \mathbf{V}^T \sigma$. The solution always exists because if there is at least one non-collinear point, the matrix is invertible. We refer to \cite{meyer2001gradient} for further technical details.

\subsubsection*{Further Diagrams}

\begin{figure*}[t!]
	\centering
	\includegraphics[scale=.22]{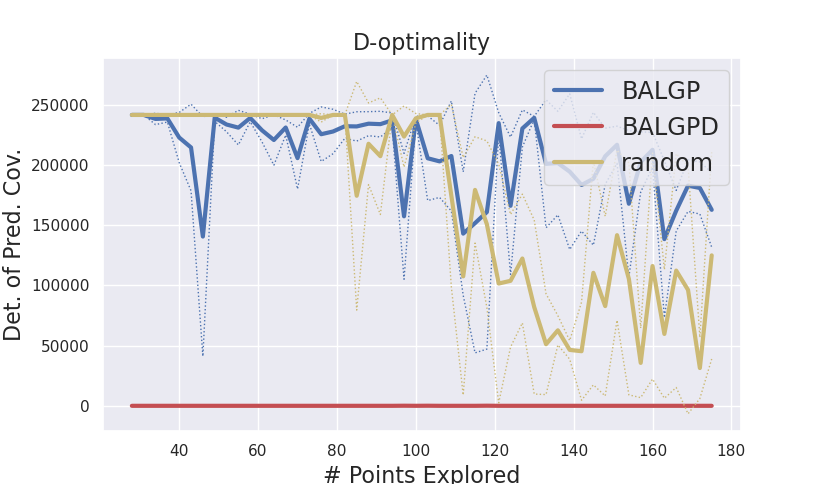}
	\includegraphics[scale=.22]{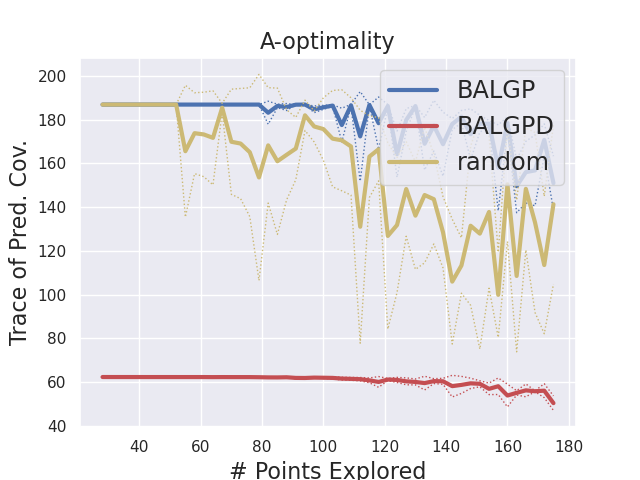}
	\includegraphics[scale=.22]{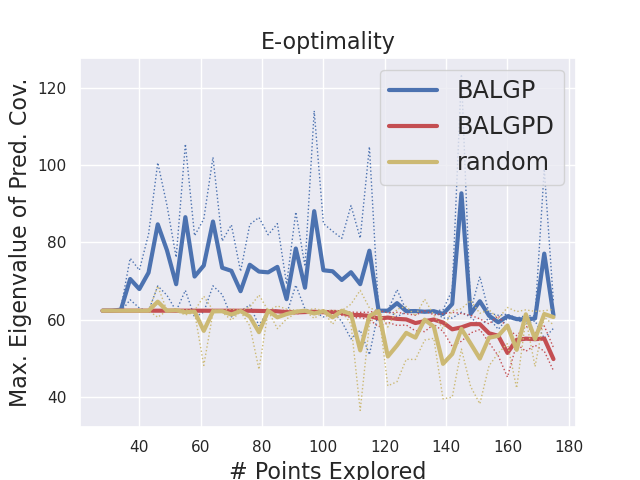}
	\caption{\emph{Map Reconstruction}: The three diagrams plot the determinant (left), trace (left) and maximum eigenvalue (right) attained under \emph{BALGPD} (red), \emph{BALGP} (blue) and random exploration (yellow), as more points are explored.}
	\label{diagram_exp3_trANDmaxeig_extra}
\end{figure*}

Again, figure \ref{diagram_exp3_trANDmaxeig_extra} presents how the determinant, trace and maximum eigenvalue of the GP predictive covariance change as more points are explored. Similarly $k(\cdot, \cdot) > 1$ at some points so it does not follow our proposition. Despite the violation of theoretical assumption in practice, from figure 6 in the main text, \emph{BALGPD} still attain a better predictive performance than \emph{BALGP}. This demonstrates the value of our approach beyond cases where theoretical assumptions are satisfied.

%
%
%
\bibliographystyle{splncs04}
\bibliography{reference}